\documentclass[10pt]{article}
\usepackage{latexsym,amsmath,amssymb,graphics,amscd, empheq}
\usepackage{multirow}
\usepackage{soul}
\usepackage{booktabs}
\usepackage{tabularx}
\usepackage[hang,footnotesize]{caption}
\usepackage[pdftex]{graphicx}
\usepackage{graphicx}
\usepackage{overpic}
\usepackage{color}
\usepackage{cancel}
\usepackage[pdftex,breaklinks,colorlinks,
  citecolor=blue,
  urlcolor=blue]{hyperref}

\addtocounter{footnote}{1}
\makeatletter
\@addtoreset{table}{bsection}
\def\thetable{\thesection.\@arabic\c@table}
\def\fps@table{h, t}
\@addtoreset{equation}{section}

\makeatother

\newtheorem{theorem}{Theorem}[section]
\newtheorem{definition}[theorem]{Definition}
\newtheorem{lemma}[theorem]{Lemma}
\newtheorem{remark}[theorem]{Remark}
\newtheorem{proposition}[theorem]{Proposition}

\newtheorem{corollary}[theorem]{Corollary}

\newcommand{\bfi}{\bfseries\itshape}
\newcommand{\vertiii}[1]{{\left\vert\kern-0.25ex\left\vert\kern-0.25ex\left\vert #1 
    \right\vert\kern-0.25ex\right\vert\kern-0.25ex\right\vert}}
\newsavebox{\savepar}

\makeatletter

\usepackage{scalerel,stackengine}
\stackMath
\newcommand\reallywidehat[1]{%
\savestack{\tmpbox}{\stretchto{%
  \scaleto{%
    \scalerel*[\widthof{\ensuremath{#1}}]{\kern-.6pt\bigwedge\kern-.6pt}%
    {\rule[-\textheight/2]{1ex}{\textheight}}
  }{\textheight}%
}{0.5ex}}%
\stackon[1pt]{#1}{\tmpbox}%
}

\begin{document}

\title{\textbf{Echo state networks are universal}}
\author{Lyudmila Grigoryeva$^{1}$ and Juan-Pablo Ortega$^{2, 3}$}
\date{}
\maketitle

\begin{abstract}
This paper shows that echo state networks are universal uniform approximants in the context of discrete-time fading memory filters with uniformly bounded inputs defined on negative infinite times. This result guarantees that any fading memory input/output system in discrete time can be realized as a  simple finite-dimensional neural network-type state-space model with a static linear readout map. This approximation is valid for infinite time intervals. The proof of this statement is based on fundamental results, also presented in this work, about the topological nature of the fading memory property and about reservoir computing systems generated by continuous reservoir maps.
\end{abstract}

\bigskip


\makeatletter
\addtocounter{footnote}{1} \footnotetext{%
Department of Mathematics and Statistics. Universit\"at Konstanz. Box 146. D-78457 Konstanz. Germany. {\texttt{Lyudmila.Grigoryeva@uni-konstanz.de} }}
\addtocounter{footnote}{1} \footnotetext{%
Universit\"at Sankt Gallen. Faculty of Mathematics and Statistics. Bodanstrasse 6.
CH-9000 Sankt Gallen. Switzerland. {\texttt{Juan-Pablo.Ortega@unisg.ch}}}
\addtocounter{footnote}{1} \footnotetext{%
Centre National de la Recherche Scientifique (CNRS). France. }
\makeatother

\medskip

\medskip

\medskip

\section{Introduction}

Many recently introduced machine learning techniques in the context of dynamical problems have much in common with system identification procedures developed in the last decades for applications in signal treatment, circuit theory and, in general, systems theory. In these problems, system knowledge is only available in the form of input-output observations and the task consists in finding or {\it learning} a model that approximates it for mainly forecasting or classification purposes. An important goal in that context is finding families of transformations that are both computationally feasible and versatile enough to reproduce a rich number of patterns just by modifying a limited number of procedural parameters. 

The versatility or flexibility of a given machine learning paradigm is usually established by proving its  {\bfi  universality}. We say that a family of transformations is universal when its elements can approximate as accurately as one wants all the elements of a  sufficiently rich class containing, for example, all continuous or even all measurable transformations. In the language of learning theory, this is equivalent to the possibility of making approximation errors arbitrarily small \cite{cucker:smale, Smale2003, cucker:zhou:book}. In more mathematical terms, the universality of a family amounts to its density in a rich class of the type mentioned above. Well-known universality results are, for example, the uniform approximation properties of feedforward neural networks established in \cite{cybenko, hornik, hornik1991} in the context of static continuous and, more generally, measurable real functions.

A first solution to this problem in the dynamic context was pioneered in the works of Fr\'echet \cite{frechet:volterra_series} and Volterra \cite{volterra:book} one century ago when they proved that finite Volterra series can be used to uniformly approximate continuous functionals defined on compact sets of continuous functions. These results were further extended in the 1950s by the MIT school lead by N. Wiener \cite{wiener:book, brilliant:volterra, george:volterra} but always under compactness assumptions on the input space and the time interval in which inputs are defined. A major breakthrough was the generalization to infinite time intervals carried out by Boyd and Chua in \cite{Boyd1985}, who formulated a uniform approximation theorem using Volterra series for operators endowed with the so called  {\bfi  fading memory property} on continuous time inputs. An input/output system is said to have fading memory when the outputs associated to inputs that are close in the recent past are close, even when those inputs may be very different in the distant past.

In this paper we address the universality or the uniform approximation problem for transformations or {\bfi  filters} of discrete time signals of infinite length that have the  fading memory property. The approximating set that we use is  generated by nonlinear state-space transformations and that is referred to as 
{\bfi reservoir computers (RC)}~\cite{jaeger2001, Jaeger04, maass1, maass2, Crook2007, verstraeten, lukosevicius} or {\bfi  reservoir systems}. These are special types of recurrent neural networks  determined by two maps, namely a {\bfi  reservoir} $F: \mathbb{R} ^N\times \mathbb{R} ^n\longrightarrow  \mathbb{R} ^N$, $n,N \in \mathbb{N} $,  and a {\bfi  readout} map $h: \mathbb{R}^N \rightarrow \mathbb{R}^d$ that under certain hypotheses transform (or filter) an infinite discrete-time input  ${\bf z}=(\ldots, {\bf z} _{-1}, {\bf z} _0, {\bf z} _1, \ldots) \in (\mathbb{R}^n) ^{\Bbb Z } $ into an output signal ${\bf y} \in (\mathbb{R} ^d)^{\Bbb Z } $ of the same type using  the state-space transformation given by:
\begin{empheq}[left={\empheqlbrace}]{align}
\mathbf{x} _t &=F(\mathbf{x}_{t-1}, {\bf z} _t),\label{reservoir equation}\\
{\bf y} _t &= h (\mathbf{x} _t), \label{readout}
\end{empheq}
where $t \in \Bbb Z $ and the dimension $N \in \mathbb{N} $ of the state vectors $\mathbf{x} _t \in \mathbb{R} ^N $ is referred to as the number of virtual {\bfi  neurons} of the system. When a RC system has a uniquely determined filter associated to it, we refer to it as the {\bfi  RC filter}.

An important advantage of the RC  approach is that, under certain hypotheses, intrinsically infinite dimensional problems regarding filters can be translated into analogous questions related to the reservoir and readout maps that generate them and that  are defined on much simpler finite dimensional spaces. This strategy has already been used in the literature in relation to the universality question in, for instance, \cite{sandberg:esn, sandberg:esn:paper, Matthews:thesis, Matthews1993, perryman:thesis, Stubberuda}. 
The universal approximation properties of feedforward neural networks~\cite{komogorovnn, arnoldnn, sprecherthesis, sprecherI, sprecherII, cybenko, hornik, hornik:derivatives, hornik1991, hornik:new:results, rueschendorf:thomsen} was used in those works to find neural networks-based families of filters that are dense in the set of approximately finite memory filters with inputs defined in the positive real half-line. Other works in connection with the universality problem in the dynamic context are~\cite{Maass2000, maass1, corticalMaass, MaassUniversality} where RC is referred to as   Liquid State Machines. In those references and in the same vein as in \cite{Boyd1985}, universal families of RC systems with inputs defined on infinite continuous time intervals were identified in the fading memory category as a corollary of the Stone-Weierstrass theorem. This approach required invoking the natural hypotheses associated to this result, like the pointwise separation property or the compactness of the input space, that was obtained as a consequence of the fading memory property. Another strand of interesting literature that we will not explore in this work has to with the Turing computability capabilities of the systems of the type that we just introduced; recent relevant works in this direction are \cite{kilian:1996, siegelmann:1997, cabessa:2015, cabessa:2016}, and references therein.

The main contribution of this paper is showing that a particularly simple type of RC systems called {\bfi echo state networks (ESNs)}  can be used as {\it  universal approximants in the context of discrete-time fading memory filters with uniformly bounded inputs defined on negative infinite times}. ESNs are RC systems of the form \eqref{reservoir equation}-\eqref{readout} given by:
\begin{empheq}[left={\empheqlbrace}]{align}
\mathbf{x} _t &=\sigma \left(A\mathbf{x}_{t-1}+ C{\bf z} _t+ \boldsymbol{\zeta}\right),\label{esn reservoir equation}\\
{\bf y} _t &= W\mathbf{x} _t. \label{esn readout}
\end{empheq}
In these equations, $C \in \mathbb{M}_{N, n} $ is called the {\bfi  input mask}, $\boldsymbol{\zeta} \in \mathbb{R} ^N $ is the {\bfi  input shift}, and $A \in \mathbb{M}_{N,N} $ is referred to as the {\bfi  reservoir matrix}.
The map $\sigma $ in  the state-space equation \eqref{esn reservoir equation} is constructed by componentwise application of a sigmoid function (like the hyperbolic tangent or the logistic function) and is called the {\bfi  activation function}. Finally, the readout map is linear in this case and implemented via the {\bfi  readout matrix} $W \in \mathbb{M}_{d, N}$. ESNs already appear in \cite{Matthews:thesis, Matthews1993} under the name of {\it recurrent networks} but it was only more recently, in the works of H. Jaeger \cite{Jaeger04}, that their outstanding performance in machine learning applications was demonstrated.

The strategy that we follow to prove that statement is a combination of what the literature refers to as {\bfi internal} and {\bfi  external approximation}. External approximation is the construction of a RC filter that approximates a given (not necessarily RC) filter. In the internal approximation problem, one is given a RC filter and builds another RC filter that approximates it by finding reservoir and readout maps that are close to those of the given one. In the external part of our proof we use a previous work \cite{RC6} where we constructed a family of RC systems with linear readouts that we called  {\bfi  non-homogeneous state affine systems (SAS)}. We showed in that paper that the RC filters associated to SAS systems uniformly approximate any discrete-time fading memory filter with uniformly bounded inputs defined on negative infinite times. Regarding the internal approximation, we  show that any RC filter, in particular SAS filters, can be approximated by ESN filters using the universal approximation property of neural networks. These two facts put together allow us to conclude that ESN filters are capable of uniformly approximating any discrete-time fading memory filter with uniformly bounded inputs. We emphasize that this result is shown exclusively for deterministic inputs using a uniform approximation criterion; an extension of this statement that accommodates stochastic inputs and $L ^p $ approximation criteria can be found in \cite{RC8}.

The paper is structured in three sections:
\begin{itemize}
\item Section \ref{Fading memory is a topological property} introduces the notation that we use all along the paper and, more importantly, specifies the topologies and Banach space structures that we need in order to talk about continuity in the context of discrete-time filters. It is worth mentioning that we characterize the fading memory property as a continuity condition of the filters that have it with respect to the product topology in the input space. On other words, {\it the fading memory property is not a metric property, as it is usually presented in the literature, but a topological one}. An important conceptual consequence of this fact is that the fading memory property does not contain any information about the rate at which systems that have it ``forget" inputs. Several corollaries can be formulated as a consequence of this fact that are very instrumental in the developments in the paper.
\item Section \ref{Internal approximation of reservoir filters} contains a collection of general results in relation with the properties of the RC systems generated by continuous reservoir maps. In particular, we provide conditions that guarantee that a unique reservoir filter can be associated to them (the so called {\bfi  echo state property}) and we  identify situations in which those filters are themselves continuous (they satisfy automatically the fading memory property). We also point out large classes of RC systems for which internal approximation is possible, that is, if the RC systems are close then so are the associated reservoir filters.
\item Section \ref{Echo state networks as universal uniform approximants} shows that {\it echo state networks are universal uniform approximants in the category of discrete-time fading memory filters with uniformly bounded inputs}.
\end{itemize}

\section{Continuous and fading memory filters}
\label{Fading memory is a topological property}

This section introduces the notation of the paper as well as general facts about filters and functionals needed in the developments that follow. The new results are contained in Section  \ref{Continuity and the fading memory property}, where we characterize the fading memory property as a continuity condition when the sequence spaces where inputs and outputs are defined are uniformly bounded and are endowed with the product topology. This feature makes this property independent of the weighting sequences that are usually introduced to define it.

\subsection{Notation}

\paragraph{Vectors and matrices.}

A column vector is denoted by a bold lower case  symbol like $\mathbf{r}$ and $\mathbf{r} ^\top $ indicates its transpose. Given a vector $\mathbf{v} \in \mathbb{R}  ^n $, we denote its entries by $v_i$, with $i \in \left\{ 1, \dots, n
\right\} $; we also write $\mathbf{v}=(v _i)_{i \in \left\{ 1, \dots, n\right\} }$.  
We denote by $\mathbb{M}_{n ,  m }$ the space of real $n\times m$ matrices with $m, n \in \mathbb{N} $. When $n=m$, we use the symbol $\mathbb{M}  _n $ to refer to the space of square matrices of order 
$n$. Given a matrix $A \in \mathbb{M}  _{n , m} $, we denote its components by $A _{ij} $ and we write $A=(A_{ij})$, with $i \in \left\{ 1, \dots, n\right\} $, $j \in \left\{ 1, \dots m\right\} $. Given a vector $\mathbf{v} \in \mathbb{R}  ^n $, the symbol $\| \mathbf{v}\|  $ stands for any norm in $\mathbb{R}  ^n $ (they are all equivalent) and is not necessarily the Euclidean one, unless it is explicitly mentioned. The open balls with respect to a given norm $\left\|\cdot \right\| $, center $\mathbf{v} \in \mathbb{R}  ^n $, and radius $r>0$ will be denoted by $B_{\left\|\cdot \right\|}(\mathbf{v}, r) $; their closures by  $\overline{B_{\left\|\cdot \right\|}(\mathbf{v}, r)} $. For any $A \in \mathbb{M}  _{n , m} $,  $\|A\| _2  $ denotes its matrix norm induced by the Euclidean norms in $\mathbb{R}^m $ and $\mathbb{R} ^n $,  and satisfies~\cite[Example 5.6.6]{horn:matrix:analysis} that $\|A\| _2=\sigma_{{\rm max}}(A)$, with $\sigma_{{\rm max}}(A)$  the largest singular value of $A$. $\|A\| _2  $ is sometimes referred to as the spectral norm of $A$. The symbol $\vertiii{\cdot}$ is reserved for the norms of operators or functionals defined on infinite dimensional spaces.

\paragraph{Sequence spaces.}
$\mathbb{N}$ denotes the set of natural numbers with the zero element included. $\Bbb Z $ (respectively, $\Bbb Z _+ $ and $\Bbb Z _- $) are the integers (respectively, the positive and the negative integers). The symbol $(\mathbb{R}^n) ^{\Bbb Z } $ denotes the set of infinite real sequences of the form ${\bf z}=(\ldots, {\bf z} _{-1}, {\bf z} _0, {\bf z} _1, \ldots) $, $ {\bf z} _i \in \mathbb{R}^n $, $i \in \Bbb Z $; $(\mathbb{R}^n) ^{\Bbb Z _ -} $ and $(\mathbb{R}^n) ^{\Bbb Z _ +} $ are the subspaces consisting of, respectively, left and right infinite sequences: $(\mathbb{R}^n) ^{\Bbb Z _ -}=\{{\bf z}=(\ldots, {\bf z} _{-2}, {\bf z} _{-1}, {\bf z} _0) \mid {\bf z} _i \in \mathbb{R}^n, i \in \mathbb{Z}_{-}\}$, $(\mathbb{R}^n) ^{\Bbb Z _ +}=\{{\bf z}=({\bf z} _0, {\bf z} _1, {\bf z} _2, \ldots) \mid {\bf z} _i \in \mathbb{R}^n, i \in \mathbb{Z}_{+}\}$.  Analogously, $(D_n) ^{\Bbb Z } $, $(D_n) ^{\Bbb Z _ -} $, and $(D_n) ^{\Bbb Z _ +} $ stand for (semi-)infinite sequences with elements in the subset $D_n\subset \mathbb{R}^n $. In most cases we  endow these infinite product spaces with the Banach space structures associated to  one of the  following two norms:
\begin{itemize}
\item The {\bfi  supremum norm}: define $\| {\bf z}\| _{\infty}:= {\rm sup}_{ t \in \Bbb Z} \left\{\| {\bf z} _t
\|\right\}$. The symbols $\ell ^{\infty}(\mathbb{R}^n) $ and $\ell_{\pm} ^{\infty}(\mathbb{R}^n) $ are used to denote the Banach spaces formed by the elements in  the corresponding infinite product spaces that have a finite supremum norm.
\item The {\bfi  weighted norm}: let $w : \mathbb{N} \longrightarrow (0,1] $ be a decreasing sequence with zero limit. We define the associated {\bfi  weighted norm } $\| \cdot \| _w $ on $(\mathbb{R}^n)^{\Bbb Z _{-}}$ associated to the {\bfi  weighting sequence} $w$ as the map:
\begin{eqnarray*}
\begin{array}{cccc}
\| \cdot \| _w :& (\mathbb{R}^n)^{\Bbb Z _{-}} & \longrightarrow & \overline{\mathbb{R}^+}\\
	&{\bf z} &\longmapsto &\| {\bf z} \| _w:= \sup_{t \in \Bbb Z_-}\{\| {\bf z}_t w_{-t}\|\}.
\end{array}
\end{eqnarray*} 
The Proposition \ref{lw is a banach space}  in Appendix \ref{lw is a banach space appendix} shows that the space
\begin{equation*}
\ell ^{w}_-({\Bbb R}^n):= \left\{{\bf z}\in \left(\mathbb{R}^n\right)^{\mathbb{Z}_{-}}\mid \| {\bf z}\| _w< \infty\right\},
\end{equation*}
endowed with weighted norm  $\| \cdot \| _w $ forms also a Banach space.
\end{itemize}
It is straightforward to show that $\left\| {\bf z} \right\|_{w}\leq \left\| {\bf z} \right\|_{\infty} $, for all $\mathbf{v} \in (\mathbb{R}^n) ^{\Bbb Z _ -} $. This implies that   $\ell_{-} ^{\infty}(\mathbb{R}^n) \subset \ell ^{w}_-({\Bbb R}^n)$ and that the inclusion map $(\ell_{-} ^{\infty}(\mathbb{R}^n), \left\| \cdot \right\|_{\infty}) \hookrightarrow (\ell ^{w}_-({\Bbb R}^n, \left\| \cdot \right\|_{w})$ is continuous.

\subsection{Filters and systems} 

\paragraph{Filters.}
Let $D_n \subset \mathbb{R}^n $ and $D_N \subset \mathbb{R}^N $. We  refer to the maps of the type $U: (D _n) ^{\Bbb Z} \longrightarrow (D_N) ^{\Bbb Z} $ as {\bfi  filters} or {\bfi  operators} and to those like $H: (D _n) ^{\Bbb Z} \longrightarrow D_N $ (or $H: (D _n) ^{\Bbb Z_\pm} \longrightarrow D_N $) as $\mathbb{R}^N $-valued  {\bfi  functionals}. These definitions will be sometimes extended to accommodate situations where the domains and the targets of the filters are not necessarily product spaces but just arbitrary subsets of $\left({\Bbb R}^n\right)^{\mathbb{Z}}  $ and $\left({\Bbb R}^N\right)^{\mathbb{Z}}  $ like, for instance, $\ell ^{\infty}(\mathbb{R}^n) $ and $\ell ^{\infty}(\mathbb{R}^N) $.

A filter $U: (D _n) ^{\Bbb Z} \longrightarrow (D_N) ^{\Bbb Z} $ is called {\bfi  causal} when for any two elements ${\bf z} , \mathbf{w} \in (D _n) ^{\Bbb Z}  $  that satisfy that ${\bf z} _\tau = \mathbf{w} _\tau$ for any $\tau \leq t  $, for a given  $t \in \Bbb Z $, we have that $U ({\bf z}) _t= U ({\bf w}) _t $. Let  $T_\tau:(D _n) ^{\Bbb Z} \longrightarrow(D _n) ^{\Bbb Z} $ be the {\bfi  time delay} operator defined by $T_\tau( {\bf z}) _t:= {\bf z}_{t- \tau}$. The filter $U$ is called {\bfi  time-invariant} (TI) when it commutes with the time delay operator, that is, $T_\tau \circ U=U  \circ T_\tau $,  for any $\tau\in \Bbb Z $ (in this expression, the two  operators $T_\tau $ have  to be understood as defined in the appropriate sequence spaces). 

We recall (see for instance~\cite{Boyd1985}) that there is a bijection between causal time-invariant filters and functionals on $(D_n)^{\Bbb Z _-} $. Indeed, consider the sets $\mathbb{F}_{(D_n)^{\mathbb{Z}_{-}}}$ and $\mathbb{H}_{(D_n)^{\mathbb{Z}_{-}}}$ defined by
\begin{eqnarray}
\mathbb{F}_{(D_n)^{\mathbb{Z}_{-}}} &:= & \left\{U: (D_n) ^{\Bbb Z} \longrightarrow (\mathbb{R}^N) ^{\Bbb Z} \mid  \mbox{$U$ is causal and time-invariant}\right\},\label{f set}\\
\mathbb{H}_{(D_n)^{\mathbb{Z}_{-}}} &:= & \left\{H: (D_n) ^{\Bbb Z_-} \longrightarrow \mathbb{R} ^N\right\}.\label{h set}
\end{eqnarray}
Then, given a time-invariant filter $U:(D_n) ^{\Bbb Z} \longrightarrow (\mathbb{R}^N) ^{\Bbb Z}$, we can associate to it a functional $H _U: (D_n) ^{\Bbb Z_-} \longrightarrow \mathbb{R} ^N$ via the assignment $H _U ({\bf z}):= U({\bf z} ^e) _0 $, where ${\bf z} ^e \in (\mathbb{R}^n)^{\Bbb Z } $ is an arbitrary extension of ${\bf z} \in (D_n)^{\Bbb Z _-} $ to $ (D_n)^{\Bbb Z } $. Let $ \boldsymbol{\Psi}: \mathbb{F}_{(D_n)^{\mathbb{Z}_{-}}} \longrightarrow \mathbb{H}_{(D_n)^{\mathbb{Z}_{-}}} $ be the map such that $\boldsymbol{\Psi}(U):= H _U $. Conversely, for any functional  $H: (D_n) ^{\Bbb Z_-} \longrightarrow \mathbb{R} ^N$, we can define a time-invariant causal filter $U_H:(D_n) ^{\Bbb Z} \longrightarrow (\mathbb{R}^N) ^{\Bbb Z}$ by $U_H({\bf z}) _t:= H((\mathbb{P}_{\Bbb Z_-} \circ T _{-t}) ({\bf z})) $, where $T _{-t} $ is the $(-t)$-time delay operator and $\mathbb{P}_{\Bbb Z_-}: (\mathbb{R}^n)^{\Bbb Z} \longrightarrow (\mathbb{R}^n)^{\Bbb Z _-} $ is the natural projection. Let $ \boldsymbol{\Phi}: \mathbb{H}_{(D_n)^{\mathbb{Z}_{-}}} \longrightarrow \mathbb{F} _{(D_n)^{\mathbb{Z}_{-}}}$ be the map such that $\boldsymbol{\Phi}(H):= U_H $. It is easy to verify that:
\begin{eqnarray*}
\boldsymbol{\Psi}\circ \boldsymbol{\Phi}&=& \mathbb{I}_{\mathbb{H}_{(D_n)^{\mathbb{Z}_{-}}}}  \quad \mbox{or, equivalently,} \quad H_{U _H}= H,\quad \mbox{for any functional} \quad H: (D_n)^{\Bbb Z _-} \rightarrow \mathbb{R}^N, \\
\boldsymbol{\Phi}\circ \boldsymbol{\Psi}&=& \mathbb{I}_{\mathbb{F}_{(D_n)^{\mathbb{Z}_{-}}}}  \quad \mbox{or, equivalently,} \quad U_{H _U} = U, \quad \mbox{for any causal TI filter} \quad U: (D_n) ^{\Bbb Z} \rightarrow (\mathbb{R}^N) ^{\Bbb Z},
\end{eqnarray*}
that is, $\boldsymbol{\Psi} $ and $\boldsymbol{\Phi}$ are inverses of each other and hence are both bijections.
Additionally, we note that the sets $\mathbb{F}_{(D_n)^{\mathbb{Z}_{-}}}  $  and $ \mathbb{H}_{(D_n)^{\mathbb{Z}_{-}}}  $ are vector spaces with naturally defined operations and that $\boldsymbol{\Psi} $ and $\boldsymbol{\Phi}$ are linear maps between them, which allows us to conclude that $\mathbb{F}_{(D_n)^{\mathbb{Z}_{-}}}  $  and $ \mathbb{H} _{(D_n)^{\mathbb{Z}_{-}}} $ are linear isomorphic. 

When a filter is causal  and time-invariant, we work in many situations just with the restriction $U: (D_n) ^{\Bbb Z_-} \longrightarrow (D_N) ^{\Bbb Z_-} $ instead of the original filter $U: (D_n) ^{\Bbb Z} \longrightarrow (D_N) ^{\Bbb Z} $ without making the distinction, since the former uniquely determines the latter. Indeed, by definition,  for any ${\bf z} \in ( D _n) ^{\Bbb Z} $ and $t \in \Bbb Z $:
\begin{equation}
\label{why we can restrict to zminus}
U ({\bf z})_t= \left(T_{-t} \left(U({\bf z})\right)\right)_0= \left(U \left(T_{-t}({\bf z})\right)\right)_0,
\end{equation}
where the second equality holds by the time-invariance of $U$ and the value in the right-hand side depends only on $\mathbb{P}_{\mathbb{Z}_{-}}\left(T_{-t}({\bf z})\right) \in (D_n) ^{\Bbb Z_-}$, by causality.

\paragraph{Reservoir systems and filters.}

Consider now the RC system determined by~\eqref{reservoir equation}--\eqref{readout} with reservoir map defined  on subsets $D _N, D' _N \subset \mathbb{R}^N $ and $D_n\subset \mathbb{R}^n $, that is, $F: D _N\times D_n\longrightarrow  D' _N$ and $h: D'_N \rightarrow \mathbb{R}^d$. There are two properties of reservoir systems that will be crucial in what follows:
\begin{itemize}
\item {\bfi Existence of solutions} property: this property holds when for each ${\bf z} \in \left(D_n\right)^{\mathbb{Z}} $  there exists an element ${\bf x} \in \left(D _N\right)^{\mathbb{Z}} $ that satisfies  the relation~\eqref{reservoir equation} for each $t \in \Bbb Z $. 
\item {\bfi  Uniqueness of solutions} or {\bfi echo state} property {\bfi  (ESP)}: it holds when the system has the existence of solutions property and, additionally, these solutions are unique.
\end{itemize}
The echo state property has deserved much attention in the context of echo state networks~\cite{jaeger2001, Jaeger04, Buehner:ESN, Yildiz2012, zhang:echo, Wainrib2016, Manjunath:Jaeger, gallicchio:esp}. We emphasize that these two properties are genuine conditions that are not automatically satisfied by all RC systems. Later on in the paper, Theorem \ref{uniform approx theorem} specifies sufficient conditions for them to hold.

The combination of the existence of solutions with the axiom of choice allows us to associate filters $U ^F: (D_n)^{\Bbb Z} \longrightarrow(D_N)^{\Bbb Z} $ to each RC system with that property via the reservoir map and~\eqref{reservoir equation}, that is, $U ^F ({\bf z}) _t := \mathbf{x} _t \in \mathbb{R} ^N $, for all $t \in \Bbb Z  $. We will denote by $U ^F _h: (D_n)^{\Bbb Z} \longrightarrow(D_d)^{\Bbb Z} $ the corresponding filter determined by the entire reservoir system, that is,  $U ^F_h ({\bf z}) _t =h \left(U ^F ({\bf z}) _t\right):= {\bf y} _t \in \mathbb{R} ^d$.  $ U ^F_h $ is said to be a {\bfi  reservoir filter} or a {\bfi  response map} associated to the RC system~\eqref{reservoir equation}--\eqref{readout}. The filters $U ^F $ and $U ^F _h $ are causal by construction. A unique reservoir filter can be associated to a reservoir system when the echo state property holds. We warn the reader that reservoir filters appear in the literature only in the presence of the ESP; that is why we sometimes make the distinction between those that come from reservoir systems that do and do not satisfy the ESP by referring to them as {\bfi   reservoir filters} and {\bfi  generalized reservoir filters}, respectively.

In the systems theory literature, the RC equations~\eqref{reservoir equation}--\eqref{readout} are referred to as the {\bfi  state-variable} or the {\bfi  internal representation} point of view and associated filters as the {\bfi   external representation} of the system.

The next proposition shows that in the presence of the ESP, reservoir filters are not only causal but also time-invariant. In that situation we can hence associate  to $U ^F _h $ a {\bfi  reservoir functional} $H^F _h : (D _n)^{\Bbb Z _-} \longrightarrow \mathbb{R}^d$ determined by $H^F _h:=H_{U ^F _h} $.

\begin{proposition}
\label{esp implies ti}
Let $D_N \subset \mathbb{R}^N $, $D_n \subset \mathbb{R}^n  $,  and $F:D_N \times D_n \longrightarrow D_N  $  be a reservoir map that satisfies the echo state property for all the elements in $\left(D_n\right)^{\mathbb{Z}} $. Then, the corresponding filter $U ^F: \left(D_n\right)^{\mathbb{Z}} \longrightarrow \left(D_N\right)^{\mathbb{Z}} $ is causal and time-invariant.
\end{proposition}

We emphasize that, as it can be seen in the proof in the appendix, it is the autonomous character of the reservoir map  that guarantees time-invariance in the previous proposition. An explicit time dependence on time in that map would spoil that conclusion.

\paragraph{Reservoir system morphisms.}

Let $N _1, N _2, n, d \in \mathbb{N} $ and let $F _1: D _{N _1}\times D_n\longrightarrow  D_{N _1}$,  $h_1: D_{N _1} \rightarrow \mathbb{R}^d$ and $F_2: D _{N _2}\times D_n\longrightarrow  D_{N _2}$,  $h_2: D_{N _2} \rightarrow \mathbb{R}^d$ be two reservoir systems. We say that a map $f:D _{N _1} \longrightarrow D_{N _2} $ is a morphism between the two systems when it satisfies the following two properties:
\begin{description}
\item [(i)]  {\bfi  Reservoir equivariance:}
$
f(F _1(\mathbf{x} _1, {\bf z}))=F _2(f(\mathbf{x}_1), {\bf z}),
$
for all $ \mathbf{x}_1 \in D _{N _1}$, and ${\bf z} \in D _n$.
\item [(ii)]  {\bfi  Readout invariance:} $h _1(\mathbf{x}_1)= h _2(f (\mathbf{x}_1)) $, for all $ \mathbf{x}_1 \in D _{N _1}$.
\end{description}
When the map $f$ has an inverse and it is also a morphism between the systems determined by the pairs $(F _2, h _2) $ and $(F _1, h _1) $ we say that $f$ is a {\bfi system isomorphism} and that the systems  $(F _1, h _1) $ and $(F _2, h _2) $ are {\bfi  isomorphic}. Given a system $F _1: D _{N _1}\times D_n\longrightarrow  D_{N _1}$,  $h_1: D_{N _1} \rightarrow \mathbb{R}^d$ and a bijection $f:D _{N _1} \longrightarrow D_{N _2} $, the map $f$ is a system isomorphism with respect to the system $F_2: D _{N _2}\times D_n\longrightarrow  D_{N _2}$,  $h_2: D_{N _2} \rightarrow \mathbb{R}^d$ defined by
\begin{eqnarray}
F_2(\mathbf{x} _2, {\bf z}) &:= & f \left(F _1(f ^{-1}(\mathbf{x} _2), {\bf z})\right), \quad \mbox{for all} \quad \mathbf{x} _2\in D _{N _2}, {\bf z} \in D _n,\label{isom system 1}\\
h_2(\mathbf{x} _2) &:= & h _1(f ^{-1}(\mathbf{x} _2))), \quad \mbox{for all} \quad \mathbf{x} _2\in D _{N _2}.\label{isom system 2}
\end{eqnarray}

The proof of the following statement is a straightforward consequence of the definitions.
\begin{proposition}
\label{morphisms consequences}
Let $F_1: D _{N _1}\times D_n\longrightarrow  D_{N _1}$,  $h_1: D_{N _1} \rightarrow \mathbb{R}^d$ and $F_2: D _{N _2}\times D_n\longrightarrow  D_{N _2}$,  $h_2: D_{N _2} \rightarrow \mathbb{R}^d$ be two reservoir systems. Let $f:D _{N _1} \longrightarrow D_{N _2} $ be a morphism between them. Then:
\begin{description}
\item [(i)]  If ${\bf x}^1 \in \left(D _{N _1}\right)^{\mathbb{Z}} $ is a solution for the reservoir map $F _1$ associated to the input ${\bf z} \in \left(D_n\right)^{\mathbb{Z}} $, then  the sequence ${\bf x}^2 \in \left(D _{N _2}\right)^{\mathbb{Z}} $ defined by ${\bf x}^2_t:= f \left({\bf x}^1 _t\right) $, $t \in \Bbb Z  $, is a solution for the reservoir map $F _2$ associated to the same input.
\item [(ii)]  If $U_{h _1}^{F _1}$ is a generalized reservoir filter for the system determined by the pair $(F _1, h _1 )$ then it is also a reservoir filter for the system $(F _2, h _2)$. Equivalently, given a generalized reservoir filter $U_{h _1}^{F _1}$   determined by $(F _1, h _1)$, there exists a generalized reservoir filter $U_{h _2}^{F _2}$ determined by $(F _2, h _2)$ such that $U_{h _1}^{F _1}= U_{h _2}^{F _2}$.
\item [(iii)] If $f$ is a system isomorphism then the  implications in the previous two points are  reversible.
\end{description}
\end{proposition}

\subsection{Continuity and the fading memory property}
\label{Continuity and the fading memory property}

In agreement with the notation introduced in the previous section, in the following paragraphs the symbol $U : \left(D_n\right)^{\mathbb{Z}_-} \longrightarrow \left(D_N\right)^{\mathbb{Z}_-} $ stands for a causal and time-invariant filter or, strictly speaking, for the restriction of $U : \left(D_n\right)^{\mathbb{Z}} \longrightarrow \left(D_N\right)^{\mathbb{Z}} $ to $\mathbb{Z}_{-} $, see  \eqref{why we can restrict to zminus};  $H _U:\left(D_n\right)^{\mathbb{Z}_-} \longrightarrow D_N $ is the associated functional, for some $D_N \subset \mathbb{R}^N $ and $D_n \subset \mathbb{R}^n  $. Analogously, $U_H $ is the filter associated to a given functional $H$.

\begin{definition}[{\bfi Continuous filters and functionals}]
\label{Continuous filters and functionals}
Let $D_N \subset \mathbb{R}^N $ and $D_n \subset \mathbb{R}^n  $ be bounded subsets such that $\left(D_n\right)^{\mathbb{Z}_-} \subset \ell ^{\infty}_-(\mathbb{R}^n) $ and $\left(D_N\right)^{\mathbb{Z}_-} \subset \ell ^{\infty}_-(\mathbb{R}^N) $. A causal and time-invariant filter $U : \left(D_n\right)^{\mathbb{Z}_-} \longrightarrow \left(D_N\right)^{\mathbb{Z}_-} $ is called {\bfi  continuous} when it is a continuous map between the metric spaces $\left(\left(D_n\right)^{\mathbb{Z}_-}, \left\|\cdot \right\|_{\infty} \right)$ and $\left(\left(D_N\right)^{\mathbb{Z}_-},  \left\|\cdot \right\|_{\infty}\right)$. An analogous prescription can be used to define {\bfi  continuous functionals} $H : \left(\left(D_n\right)^{\mathbb{Z}_-}, \left\|\cdot \right\|_{\infty} \right) \longrightarrow \left(D_N,  \left\|\cdot \right\|\right) $. 
\end{definition}

The following proposition shows that when filters are causal and time-invariant, their continuity can be read out of their corresponding functionals and viceversa. 

\begin{proposition}
\label{continuous functional iff filter}
Let $D_n \subset \mathbb{R}^n $ and $D_N \subset \mathbb{R}^N  $ be such that $\left(D_n\right)^{\mathbb{Z}_-} \subset \ell ^{\infty}_-(\mathbb{R}^n) $ and $\left(D_N\right)^{\mathbb{Z}_-} \subset \ell ^{\infty}_-(\mathbb{R}^N) $. Let $U : \left(D_n\right)^{\mathbb{Z}_-} \longrightarrow \left(D_N\right)^{\mathbb{Z}_-} $ be a causal and time-invariant filter, $H : \left(D_n\right)^{\mathbb{Z}_-} \longrightarrow D_N$ a functional,  and let $\boldsymbol{\Phi} $ and $\boldsymbol{\Psi} $ be the maps defined in the previous section. Then, if the filter $U$ is continuous then so is the associated functional $\boldsymbol{\Psi} (U) =:H _U  $. Conversely, if $H$ is continuous then so is $\boldsymbol{\Phi} (H) =:U_H  $.
Define now the vector spaces
\begin{eqnarray}
\mathbb{F}_{(D_n)^{\mathbb{Z}_{-}}} ^{\infty}&:= & \left\{U: (D_n) ^{\Bbb Z_-} \longrightarrow \ell ^{\infty}_-(\mathbb{R}^N) \mid  \mbox{$U$ is causal, time-invariant, and continuous}\right\},\label{f set continuous}\\
\mathbb{H}_{(D_n)^{\mathbb{Z}_{-}}} ^{\infty}&:= & \left\{H: (D_n) ^{\Bbb Z_-} \longrightarrow \mathbb{R} ^N \mid  \mbox{$H$ is continuous}\right\}.\label{h set continuous}
\end{eqnarray}
The previous statements guarantee that the maps $\boldsymbol{\Psi} $ and $\boldsymbol{\Phi} $ restrict to the maps (that we denote with the same symbol)
$\boldsymbol{\Psi}:\mathbb{F}_{(D_n)^{\mathbb{Z}_{-}}} ^{\infty} \longrightarrow \mathbb{H}_{(D_n)^{\mathbb{Z}_{-}}} ^{\infty}$ and $\boldsymbol{\Phi}:\mathbb{H}_{(D_n)^{\mathbb{Z}_{-}}} ^{\infty} \longrightarrow \mathbb{F}_{(D_n)^{\mathbb{Z}_{-}}} ^{\infty}$ that are linear isomorphisms and are inverses of each other.
\end{proposition}

\begin{definition}[{\bfi Fading memory filters and functionals}]
\label{Fading memory filters and functionals}
Let $w : \mathbb{N} \longrightarrow (0,1] $ be a weighting sequence and let $D_N \subset \mathbb{R}^N $ and $D_n \subset \mathbb{R}^n  $ be such that $\left(D_n\right)^{\mathbb{Z}_-} \subset \ell ^{w}_-({\Bbb R}^n)$ and $\left(D_N\right)^{\mathbb{Z}_-} \subset \ell ^{w}_-(\mathbb{R}^N) $. We say that a causal and time-invariant filter $U : \left(D_n\right)^{\mathbb{Z}_-} \longrightarrow \left(D_N\right)^{\mathbb{Z}_-} $ (respectively, a functional $H : \left(D_n\right)^{\mathbb{Z}_-} \longrightarrow D_N $) satisfies the {\bfi  fading memory property (FMP)} with respect to the sequence $w$ when it is a continuous map between the metric spaces $\left(\left(D_n\right)^{\mathbb{Z}_-}, \left\|\cdot \right\|_w \right)$ and $\left(\left(D_N\right)^{\mathbb{Z}_-},  \left\|\cdot \right\|_{w}\right)$ (respectively, $\left(\left(D_n\right)^{\mathbb{Z}_-}, \left\|\cdot \right\|_w \right)$ and $\left(D_N,  \left\|\cdot \right\|\right)$). 
If the weighting sequence $w$ is such that $w _t= \lambda ^t $, for some $\lambda\in (0,1) $ and all $t \in \mathbb{N}  $,  then $U$ is said to have the $\lambda $-{\bfi exponential fading memory property}. We define the sets
\begin{eqnarray}
\!\!\!\!\!\!\!\!\!\!\!\!\mathbb{F}_{(D_n)^{\mathbb{Z}_{-}},(D_N)^{\mathbb{Z}_{-}}} ^{w}&:= &\left\{U: (D_n) ^{\Bbb Z_-} \longrightarrow (D_N)^{\mathbb{Z}_{-}} \mid  \mbox{$U$ causal, time-invariant, and FMP w.r.t. $w$}\right\},\label{f set fmp 1}\\
\!\!\!\!\!\!\!\!\!\!\!\!\mathbb{H}_{(D_n)^{\mathbb{Z}_{-}},(D_N)^{\mathbb{Z}_{-}}} ^{w}&:= &\left\{H: (D_n) ^{\Bbb Z_-} \longrightarrow D_N \mid  \mbox{$H$ is FMP with respect to $w$}\right\}.\label{h set fmp 1}
\end{eqnarray}
These definitions can be extended by replacing the product set $\left(D_N\right)^{\mathbb{Z}_-} $ by any subset of $\ell ^{w}_-(\mathbb{R}^N)$ that is not necessarily a product space. In particular, we define the sets
\begin{eqnarray}
\mathbb{F}_{(D_n)^{\mathbb{Z}_{-}}} ^{w}&:= & \left\{U: (D_n) ^{\Bbb Z_-} \longrightarrow \ell ^{w}_-(\mathbb{R}^N) \mid  \mbox{$U$ is causal, time-invariant, and FMP w.r.t. $w$}\right\},\label{f set fmp}\\
\mathbb{H}_{(D_n)^{\mathbb{Z}_{-}}} ^{w}&:= & \left\{H: (D_n) ^{\Bbb Z_-} \longrightarrow \mathbb{R} ^N \mid  \mbox{$H$ is FMP with respect to $w$}\right\}.\label{h set fmp}
\end{eqnarray}
\end{definition}

Definitions \ref{Continuous filters and functionals} and \ref{Fading memory filters and functionals} can be easily reformulated in terms of more familiar $\epsilon$-$\delta $-type criteria, as they were introduced in \cite{Boyd1985}. For example, the continuity of the functional $H : \left(D_n\right)^{\mathbb{Z}_-} \longrightarrow D_N $ is equivalent  to stating that  for any ${\bf z} \in (D_n)^{\Bbb Z _{-}} $ and any $\epsilon>0  $, there exists a $\delta(\epsilon)> 0 $ such that for any ${\bf s} \in (D_n)^{\Bbb Z _{-}}$ that satisfies that
\begin{equation}
\label{continuity epsilon delta}
\| {\bf z} - {\bf s}\|_{\infty}=\sup_{t \in \Bbb Z_-}\{\| {\bf z}_t-{\bf s}_t \|\}< \delta(\epsilon), \quad \mbox{then} \quad \|H _U({\bf z})-H _U({\bf s})\|< \epsilon.
\end{equation}
Regarding the fading memory property, it suffices to replace the implication in \eqref{continuity epsilon delta} by
\begin{equation}
\label{fmp epsilon delta}
\| {\bf z} - {\bf s}\|_{w}=\sup_{t \in \Bbb Z_-}\{\| {\bf z}_t-{\bf s}_t \|w_{-t}\}< \delta(\epsilon), \quad \mbox{then} \quad \|H _U({\bf z})-H _U({\bf s})\|< \epsilon.
\end{equation}

A very important part of the results that follow concern {\bfi  uniformly bounded}  families of sequences, that is, subsets of $\left({\Bbb R}^n\right)^{\mathbb{Z}_{-}} $ of the form
\begin{equation}
\label{Kset}
K_{M}:=\left\{ {\bf z} \in \left({\Bbb R}^n\right)^{\mathbb{Z}_{-}} \mid \| {\bf z}_t\| \leq M \quad \mbox{for all} \quad t \in \Bbb Z _{-} \right\}, \quad \mbox{for some $M>0 $.}
\end{equation}
It is straightforward to show that $K _M\subset\ell_{-} ^{\infty}(\mathbb{R}^n) \subset \ell ^{w}_-({\Bbb R}^n)$, for all $M>0 $ and any weighting sequence $w$. A very useful fact is that the relative topology induced by $(\ell ^{w}_-({\Bbb R}^n), \left\|\cdot \right\|_w) $ in $K _M $ coincides with the one induced by the product topology in $\left({\Bbb R}^n\right)^{\mathbb{Z}_{-}} $. This is a consequence of the following result that is a slight generalization of \cite[Theorem 20.5]{Munkres:topology}. A proof is provided in Appendix \ref{proof of product topology for uniformly bounded} for the sake of completeness.

\begin{theorem}
\label{product topology for uniformly bounded}
Let $\left\|\cdot \right\|: \mathbb{R}^n \longrightarrow [0, \infty) $ be a norm in ${\Bbb R}^n$, $M>0 $, and let $w : \mathbb{N} \longrightarrow (0,1] $ be a weighting sequence. Let $\overline{d} _M(\mathbf{a},\mathbf{b}):=\min \left\{\left\|\mathbf{a}-\mathbf{b}\right\|, M\right\}$, $\mathbf{a} ,\mathbf{b} \in {\Bbb R}^n $, be a bounded metric on ${\Bbb R}^n $ and define the $w$-{\bfi weighted metric} $D_w^M  $ on $({\Bbb R}^n)^{\mathbb{Z}_{-}} $ as
\begin{equation}
\label{def of weighted metric}
D_w^M(\mathbf{x}, {\bf y}):=\sup_{t \in \mathbb{Z}_{-}} \left\{\overline{d} _M(\mathbf{x}_t, {\bf y}_t)w_{-t}\right\}, \quad \mathbf{x}, {\bf y} \in ({\Bbb R}^n)^{\mathbb{Z}_{-}}.
\end{equation}   
Then $D_w^M$ is a metric  that induces the product topology on $({\Bbb R}^n)^{\mathbb{Z}_{-}} $. The space $({\Bbb R}^n)^{\mathbb{Z}_{-}} $ is complete relative to this metric.
\end{theorem}

An important consequence that can be drawn from this theorem is that all the weighted norms induce the same topology on the subspaces formed by uniformly bounded sequences. An obvious consequence of this fact is that continuity  with respect to this topology can be defined without the help of weighting sequences or, equivalently, filters or functionals with uniformly bounded inputs that have the fading memory with respect to a weighting sequence, have the same feature with respect to any other weighting sequence. We make this more specific in the following statements.

\begin{corollary}
\label{all weighted norms are the same}
Let $M>0 $ and let $K_{M}:=\left\{ {\bf z} \in \left({\Bbb R}^n\right)^{\mathbb{Z}_{-}} \mid \| {\bf z}_t\| \leq M \quad \mbox{for all} \quad t \in \Bbb Z _{-} \right\} $ be a subset of $\left({\Bbb R}^n\right)^{\mathbb{Z}_{-}} $ formed by uniformly bounded sequences. Let $w : \mathbb{N} \longrightarrow (0,1] $ be an arbitrary weighting sequence. Then, the metric induced by the weighted norm $\left\|\cdot \right\|_w  $ on $K _M $ coincides with $D_w^{2M} $. Moreover, since $D_w^{2M} $ induces the product topology on $K _M=\left(\overline{B_{\left\|\cdot \right\|}(\mathbf{0}, M)}\right)^{\mathbb{Z}_{-}} $, we can conclude that all the weighted norms induce the same topology on $K _M $. We recall that $\overline{B_{\left\|\cdot \right\|}(\mathbf{0}, M)} $ is the closure of the ball with radius $M$ centered at the origin, with respect to the norm $\left\|\cdot \right\|$ in ${\Bbb R}^n$. The same conclusion holds when instead of $K _M $ we consider the set $(D_n) ^{\mathbb{Z}_{-}} $, with $D_n $ a compact subset of ${\Bbb R}^n $.
\end{corollary}

Theorem \ref{product topology for uniformly bounded} can also be used to give a quick alternative proof in discrete time to an important compactness result originally formulated in Boyd and Chua in \cite[Lemma 1]{Boyd1985} for continuous time and, later on, in \cite{RC6} for discrete time. The next corollary contains an additional completeness statement.

\begin{corollary}
\label{km compact complete}
Let $K _M$ be the set of uniformly bounded sequences, defined as in \eqref{Kset}, and let  $w : \mathbb{N} \longrightarrow (0,1] $ be a weighting sequence. Then, $ \left(K _M, \left\|\cdot \right\|_w\right) $ is a compact, complete, and convex subset of the Banach space $(\ell ^{w}_-({\Bbb R}^n), \left\|\cdot \right\|_w) $. The compactness and the completeness statements also hold when instead of $K _M $ we consider the set $(D_n) ^{\mathbb{Z}_{-}} $, with $D_n $ a compact subset of ${\Bbb R}^n $; if $D_n $ is additionally convex then the convexity of $(D_n) ^{\mathbb{Z}_{-}} $ is also guaranteed.
\end{corollary}

It is important to point out that the coincidence between the product topology and the topologies induced by weighted norms that we described in Corollary \ref{all weighted norms are the same} only occurs for uniformly bounded sets of the type introduced in \eqref{Kset}. As we state in the next result, the norm topology in $\ell ^{w}_-({\Bbb R}^n) $ is strictly finer than the one induced by the product topology in $ \left(\mathbb{R}^n\right) ^{\mathbb{Z}_-} $.

\begin{proposition}
\label{in lww norm finer than product}
Let  $w : \mathbb{N} \longrightarrow (0,1] $ be a weighting sequence and let $(\ell ^{w}_-({\Bbb R}^n), \left\|\cdot \right\|_w) $ be the Banach space constructed using the corresponding weighted norm on the space of left infinite sequences with elements in $\mathbb{R}^n $. The norm topology in $\ell ^{w}_-({\Bbb R}^n) $ is strictly finer than the subspace topology induced by the product topology in $\left(\mathbb{R}^n\right)^{\mathbb{Z}_{-}} $ on $\ell ^{w}_-({\Bbb R}^n) \subset \left(\mathbb{R}^n\right)^{\mathbb{Z}_{-}}$.
\end{proposition}

The results that we just proved imply an elementary property of the sets that we defined in \eqref{f set fmp 1}-\eqref{h set fmp 1} and \eqref{f set fmp}-\eqref{h set fmp} that we state in the following lemma.

\begin{lemma}
\label{fs and ^sfor w}
Let $M>0$  and let $w $ be a weighting sequence. Let $U:K _M \longrightarrow \ell ^{w}_-(\mathbb{R}^N)  $  (respectively, $H: K _M \longrightarrow \mathbb{R}^N $) be and element of $\mathbb{F}_{K _M} ^{w} $  (respectively, $\mathbb{H}_{K _M} ^{w} $). Then there exists $L>0 $ such that $U(K _M) \subset K _L $ (respectively, $H(K_M)\subset \overline{B_{\left\|\cdot \right\|}(\mathbf{0}, L)}) $) and we can hence conclude that $U \in \mathbb{F}_{K _M, K _L} ^{w} $ (respectively, $H \in \mathbb{H}_{K _M, K _L} ^{w} $). Conversely, the inclusion $\mathbb{F}_{K _M, K _L} ^{w} \subset \mathbb{F}_{K _M} ^{w} $ (respectively, $\mathbb{H}_{K _M, K _L} ^{w} \subset \mathbb{H}_{K _M} ^{w} $) holds true for any $M>0 $. The sets $\mathbb{F}_{K _M} ^{w} $ and $\mathbb{H}_{K _M} ^{w} $ are vector spaces.
\end{lemma}

The next proposition spells out how the fading memory property is independent of the weighting sequence that is used to define it, which shows its intrinsically topological nature. A conceptual consequence of this fact is that the fading memory property does not contain any information about the rate at which systems that have it ``forget" inputs.  A similar statement in the continuous time setup has been formulated in \cite{sandberg:fmp}. Additionally, there is a bijection between FMP filters and functionals.
 
\begin{proposition}
\label{FMP independent of w}
Let $K _M \subset  \left({\Bbb R}^n\right)^{\mathbb{Z}_{-}}$ and $K _L \subset  \left({\Bbb R}^N\right)^{\mathbb{Z}_{-}} $ be subsets of uniformly bounded sequences defined as in \eqref{Kset} and let  $w : \mathbb{N} \longrightarrow (0,1] $ be a weighting sequence. Let  $U: K _M \longrightarrow K _L$ be a causal and time-invariant  filter and  let $H:K _M \longrightarrow \overline{B_{\left\|\cdot \right\|}( {\bf 0},L)}$ be a functional. Then: 
\begin{description}
\item [(i)] If $U$ (respectively $H$) has the fading memory property with respect to the weighting sequence $w$, then it has the same property with respect to any other weighting sequence. In particular, this implies that
\begin{equation*}
\mathbb{F}_{K _M, K _L} ^{w}=\mathbb{F}_{K _M, K _L} ^{w'}\quad \mbox{and} \quad\mathbb{H}_{K _M, K _L} ^{w}=\mathbb{H}_{K _M, K _L} ^{w'}, \quad \mbox{for any weighting sequence $w'$.}
\end{equation*}
In what follows we just say that $U$ (respectively $H$) has the fading memory property and denote
\begin{equation*}
\mathbb{F}_{K _M, K _L} ^{{\rm FMP}}:=\mathbb{F}_{K _M, K _L} ^{w}\quad \mbox{and} \quad\mathbb{H}_{K _M, K _L} ^{{\rm FMP}}:=\mathbb{H}_{K _M, K _L} ^{w}, \quad \mbox{for any weighting sequence $w$.}
\end{equation*}
The same statement holds true for the vector spaces $\mathbb{F}_{K _M } ^{w} $ and $\mathbb{H}_{K _M } ^{w} $, that will be denoted in the sequel by $\mathbb{F}_{K _M } ^{{\rm FMP}} $ and $\mathbb{H}_{K _M } ^{{\rm FMP}} $, respectively.
\item [(ii)] Let $\boldsymbol{\Phi} $ and $\boldsymbol{\Psi} $ be the maps defined in the previous section. Then, if the filter $U$ has the fading memory property then so does the associated functional $\boldsymbol{\Psi} (U) =:H _U  $. Analogously, if $H$ has the fading memory property, then so does $\boldsymbol{\Phi} (H) =:U_H  $. This implies that the maps $\boldsymbol{\Psi} $ and $\boldsymbol{\Phi} $ restrict to maps (that we denote with the same symbols) $\boldsymbol{\Psi}:\mathbb{F}_{K _M, K _L} ^{{\rm FMP}} \longrightarrow \mathbb{H}_{K _M, K _L} ^{{\rm FMP}}$ and $\boldsymbol{\Phi}:\mathbb{H}_{K _M, K _L} ^{{\rm FMP}} \longrightarrow \mathbb{F}_{K _M, K _L} ^{{\rm FMP}}$ that are inverses of each other. The same applies to $\boldsymbol{\Psi}:\mathbb{F}_{K _M} ^{{\rm FMP}} \longrightarrow \mathbb{H}_{K _M} ^{{\rm FMP}}$ and $\boldsymbol{\Phi}:\mathbb{H}_{K _M} ^{{\rm FMP}} \longrightarrow \mathbb{F}_{K _M} ^{{\rm FMP}}$ that, in this case, are  linear isomorphisms.
\end{description}
The same statements can be formulated  when instead of $K _M $ and $K _L $ we consider the sets $(D_n) ^{\mathbb{Z}_{-}} $ and $(D_N) ^{\mathbb{Z}_{-}} $, with $D_n $ and $D_N $ compact subsets of ${\Bbb R}^n $ and $\mathbb{R}^N $, respectively.
\end{proposition}

In the conditions of the previous proposition, the vector spaces $\mathbb{F}_{K _M} ^{{\rm FMP}} $ and  $ \mathbb{H}_{K _M}^{{\rm FMP}}$ can be endowed with a norm. More specifically, let $U: K _M \longrightarrow \ell ^{w}_-({\Bbb R}^n)$ be a   filter and let $H:K _M \longrightarrow \mathbb{R}^N$ be a functional that have the FMP. Define:
\begin{eqnarray}
\vertiii{U}_{\infty} &:=&\sup_{{\bf z} \in K _M} \left\{\left\|U ({\bf z})\right\|_ \infty\right\}=\sup_{{\bf z} \in K _M} \left\{\sup_{t \in \mathbb{Z}_{-}}\left\{\left\|U ({\bf z})_t\right\|\right\}\right\},\label{norm of U}\\
\vertiii{H}_{\infty} &:=&\sup_{{\bf z} \in K _M} \left\{\left\|H ({\bf z})\right\|\right\}.\label{norm of H}
\end{eqnarray}
The compactness of $(K _M, \left\|\cdot \right\|_w) $ guaranteed by Corollary \ref{km compact complete} and the fact that by Lemma \ref{fs and ^sfor w} $U$ and $H$ map into uniformly bounded sequences and a compact subspace of $\mathbb{R}^N  $, respectively,   ensures that  the values in \eqref{norm of U} and \eqref{norm of H} are finite, which makes $\left(\mathbb{F}_{K _M} ^{{\rm FMP}},  \vertiii{\cdot }_{\infty} \right)$ and  $ \left(\mathbb{H}^{{\rm FMP}}_{K _M}, \vertiii{\cdot }_{\infty}  \right)$  into normed spaces that, as we will see in the next result, are linearly homeomorphic. For any $L>0 $ these norms restrict to the spaces $\mathbb{F}_{K _M, K _L} ^{{\rm FMP}} $ and $\mathbb{H}_{K _M, K _L} ^{{\rm FMP}} $, which are in general not linear but become nevertheless metric spaces. 

\begin{proposition}
\label{linear homeomorphism prop}
The linear isomorphism $\boldsymbol{\Psi}: \left(\mathbb{F}_{K _M} ^{{\rm FMP}},  \vertiii{\cdot }_{\infty} \right)\longrightarrow \left(\mathbb{H}_{K _M}^{{\rm FMP}}, \vertiii{\cdot }_{\infty}  \right)$ and its inverse $\boldsymbol{\Phi} $ satisfy that
\begin{eqnarray}
\vertiii{\boldsymbol{\Psi}(U )}_{\infty}&\leq&
\vertiii{U}_ \infty, \quad \mbox{for any} \quad U \in  \mathbb{F}_{K _M} ^{{\rm FMP}}, \label{first ineq psis}\\
\vertiii{\boldsymbol{\Phi}(H)}_{\infty}&\leq& \vertiii{H}_{\infty}, \quad \mbox{for any} \quad H \in \mathbb{H}_{K _M}^{{\rm FMP}}.\label{second ineq psis}
\end{eqnarray}
These inequalities imply that these two maps are continuous linear bijections and hence the spaces $\left(\mathbb{F}_{K _M} ^{{\rm FMP}},  \vertiii{\cdot }_{\infty} \right)$ and  $ \left(\mathbb{H}_{K _M}^{{\rm FMP}}, \vertiii{\cdot }_{\infty}  \right)$ are linearly homeomorphic. Equivalently, the following  diagram commutes and all the maps in it are linear and continuous
$$\minCDarrowwidth55pt
\begin{CD}
\left(\mathbb{F}_{K _M} ^{{\rm FMP}},  \vertiii{\cdot }_{\infty} \right)     @>\boldsymbol{\Psi}>>  \left(\mathbb{H}_{K _M}^{{\rm FMP}}, \vertiii{\cdot }_{\infty}  \right)\\
@A{\rm Id}_{\mathbb{F}_{K _M} ^{{\rm FMP}}}AA        @VV{\rm Id}_{\mathbb{H}_{K _M} ^{{\rm FMP}}}V\\
\left(\mathbb{F}_{K _M} ^{{\rm FMP}},  \vertiii{\cdot }_{\infty} \right)     @<\boldsymbol{\Phi}<<  \left(\mathbb{H}_{K _M}^{{\rm FMP}}, \vertiii{\cdot }_{\infty}  \right).
\end{CD}$$
For any $L>0 $, the inclusions $\left(\mathbb{F}_{K _M, K _L} ^{{\rm FMP}},  \vertiii{\cdot }_{\infty} \right) \hookrightarrow \left(\mathbb{F}_{K _M} ^{{\rm FMP}},  \vertiii{\cdot }_{\infty} \right)$ and $\left(\mathbb{H}_{K _M, K _L} ^{{\rm FMP}},  \vertiii{\cdot }_{\infty} \right)  \hookrightarrow \left(\mathbb{H}_{K _M} ^{{\rm FMP}},  \vertiii{\cdot }_{\infty} \right)$ (see Lemma \ref{fs and ^sfor w}) are continuous and so are the restricted bijections (that we denote with the same symbols) $\boldsymbol{\Psi}:(\mathbb{F}_{K _M, K _L} ^{{\rm FMP}},  \vertiii{\cdot }_{\infty}) \longrightarrow (\mathbb{H}_{K _M, K _L} ^{{\rm FMP}},  \vertiii{\cdot }_{\infty}) $ and $\boldsymbol{\Phi}:(\mathbb{H}_{K _M, K _L} ^{{\rm FMP}},  \vertiii{\cdot }_{\infty})  \longrightarrow (\mathbb{F}_{K _M, K _L} ^{{\rm FMP}},  \vertiii{\cdot }_{\infty}) $ that are inverses of each other. The last statement is a consequence of the following inequalities:
\begin{eqnarray}
\vertiii{\boldsymbol{\Psi}(U _1 )-\boldsymbol{\Psi}(U _2)}_{\infty}&\leq&
\vertiii{U_1- U _2}_ \infty, \quad \mbox{for any} \quad U_1, U _2 \in  \mathbb{F}_{K _M, K _L} ^{{\rm FMP}}, \label{first ineq psis kl}\\
\vertiii{\boldsymbol{\Phi}(H_1)-\boldsymbol{\Phi}(H_2)}_{\infty}&\leq& \vertiii{H_1- H _2}_{\infty}, \quad \mbox{for any} \quad H _1, H _2 \in \mathbb{H}_{K _M, K _L}^{{\rm FMP}}.\label{second ineq psis kl}
\end{eqnarray}
The same statements can be formulated  when instead of $K _M $ and $K _L $ we consider the sets $(D_n) ^{\mathbb{Z}_{-}} $ and $(D_N) ^{\mathbb{Z}_{-}} $, with $D_n $  and $D_N $  compact subsets of ${\Bbb R}^n $ and ${\Bbb R}^N$, respectively.
\end{proposition}

\section{Internal approximation of reservoir filters}
\label{Internal approximation of reservoir filters}

This section characterizes situations under which reservoir filters can be uniformly approximated by finding uniform approximants for the corresponding reservoir systems. Such a statement is part of the next theorem that also identifies criteria for the availability of the echo state and the fading memory properties (recall that we used the acronyms ESP and FMP, respectively). As it was already mentioned, a reservoir system has the ESP when it has a unique   semi-infinite solution for each semi-infinite input. We also recall that in the presence of uniformly bounded inputs, as it was shown in Section \ref{Continuity and the fading memory property}, the FMP amounts to the continuity of a reservoir filter with respect to the product topologies on the input and output spaces. The completeness and compactness of those spaces established in Corollary \ref{km compact complete} allows us to use various fixed point theorems to show that solutions for reservoir systems exist under very weak hypotheses and that for contracting and continuous reservoir maps (we define this below) these solutions are unique and depend continuously on the inputs. Said differently, {\it contracting continuous reservoir maps  induce reservoir filters that automatically have the echo state and the fading memory properties}.

\begin{theorem}
\label{uniform approx theorem}
Let $K _M \subset  \left({\Bbb R}^n\right)^{\mathbb{Z}_{-}}$ and $K _L \subset  \left({\Bbb R}^N\right)^{\mathbb{Z}_{-}} $ be subsets of uniformly bounded sequences defined as in \eqref{Kset} and let $F: \overline{B_{\left\|\cdot \right\|}({\bf 0}, L)} \times \overline{B_{\left\|\cdot \right\|}({\bf 0}, M)} \longrightarrow \overline{B_{\left\|\cdot \right\|}({\bf 0}, L)} $ be a continuous reservoir map.
\begin{description}
\item [(i)] {\bf Existence of solutions:} for each ${\bf z} \in K _M$ there exists a $\mathbf{x} \in K _L$ (not necessarily unique) that solves the reservoir equation associated to $F$, that is,
\begin{equation*}
\mathbf{x}_t=F( \mathbf{x}_{t-1}, {\bf z}_t), \quad \mbox{for all $t \in \mathbb{Z}_{-}$.} 
\end{equation*}
\item [(ii)] {\bf Uniqueness and continuity of solutions (ESP and FMP):} suppose that the reservoir map $F$ is a contraction, that is, there exists $0<r<1$ such that  for all $\mathbf{u}, \mathbf{v} \in \overline{B_{\left\|\cdot \right\|}({\bf 0}, L)}$, $\mathbf{z} \in \overline{B_{\left\|\cdot \right\|}({\bf 0}, M)}$, one has 
\begin{equation*}
\left\|F(\mathbf{u}, {\bf z})-F(\mathbf{v}, {\bf z})\right\|\leq r \left\|\mathbf{u}- \mathbf{v}\right\|.
\end{equation*}
Then, the reservoir system associated to $F$ has the echo state property. Moreover, this system has a unique associated causal and time-invariant filter $U _F:K _M \longrightarrow K _L $ that has the fading memory property, that is, $U _F \in \mathbb{F}_{K _M, K _L} ^{{\rm FMP}} $. The set $U _F (K _M)$ of accessible states  of the filter $U _F  $ is compact.
\item [(iii)] {\bf Internal approximation property:} 
let $F _1, F _2:\overline{B_{\left\|\cdot \right\|}({\bf 0}, L)} \times \overline{B_{\left\|\cdot \right\|}({\bf 0}, M)} \longrightarrow \overline{B_{\left\|\cdot \right\|}({\bf 0}, L)}  $ be two continuous reservoir maps such that  $F _1  $ is a contraction with constant $0<r<1$ and $F _2 $ has the existence of solutions property. Let $U _{F _1}, U _{F _2}:K _M \longrightarrow K _L $ be the corresponding filters (if $F _2 $ does not have the ESP, then $U _{F _2} $ is just a generalized filter). Then, for any  $\epsilon>0 $, we have that 
\begin{equation}
\label{uniform mathema statement}
\left\|F _1-F _2\right\|_{\infty}< \delta(\epsilon):=(1-r) \epsilon \quad \mbox{implies that} \quad
\vertiii{U_{F _1}-U_{F _2}}_{\infty}< \epsilon.
\end{equation}
\end{description}
Part {\bf (i)} also holds true  when instead of $K _M $ and $K _L $ we consider the sets $(D_n) ^{\mathbb{Z}_{-}} $ and $(D_N) ^{\mathbb{Z}_{-}} $, with $D_n $ and $D_N $ compact and convex subsets of ${\Bbb R}^n $ and $\mathbb{R}^N $, respectively, that replace the closed balls $\overline{B_{\left\|\cdot \right\|}({\bf 0}, M)} $ and $\overline{B_{\left\|\cdot \right\|}({\bf 0}, L)} $. The same applies to parts {\bf (ii)} and {\bf (iii)} but, this time, the convexity hypothesis is not needed.
\end{theorem}

\noindent  Define the set $\mathbb{K}_{K _M, K _L}:= \left\{F: \overline{B_{\left\|\cdot \right\|}({\bf 0}, L)} \times \overline{B_{\left\|\cdot \right\|}({\bf 0}, M)} \longrightarrow \overline{B_{\left\|\cdot \right\|}({\bf 0}, L)}\mid \mbox{$F$ is a continuous contraction} \right\} $. Using the notation introduced in the previous section, the statement in \eqref{uniform mathema statement} and part {\bf (ii)} of the theorem automatically imply that the map
\begin{equation*}
\begin{array}{cccc}
\Xi : &(\mathbb{K}_{K _M, K _L}, \left\| \cdot \right\|_{\infty})& \longrightarrow &\left(\mathbb{F}_{K _M, K _L} ^{{\rm FMP}},  \vertiii{\cdot }_{\infty} \right)\\
	& F&\longmapsto &U _F
\end{array}
\end{equation*}
is continuous and by Proposition \ref{linear homeomorphism prop}, the map that associates to each $F \in \mathbb{K}_{K _M, K _L}$ the corresponding functional $H _F $, that is,
\begin{equation*}
\begin{array}{cccc}
\boldsymbol{\Psi} \circ \Xi : &(\mathbb{K}_{K _M, K _L}, \left\| \cdot \right\|_{\infty})& \longrightarrow &\left(\mathbb{H}_{K _M, K _L} ^{{\rm FMP}},  \vertiii{\cdot }_{\infty} \right)\\
	& F&\longmapsto &H _F,
\end{array}
\end{equation*}
is also continuous.

\medskip

\noindent\textbf{Proof of the theorem. \ \ }{\bf (i)} We start by defining, for each ${\bf z} \in K _M$, the map given by
\begin{equation*}
\begin{array}{cccc}
\mathcal{F}_{\bf z}: & K _L &\longrightarrow &K _L\\
	&\mathbf{x}&\longmapsto & \left(\mathcal{F}_{\bf z}(\mathbf{x})\right)_t:=F(\mathbf{x} _{t-1}, {\bf z}_t).
\end{array}
\end{equation*}
We show first that $\mathcal{F}_{\bf z} $ can be written as a product of continuous functions. Indeed:
\begin{equation}
\label{product decomposition}
\mathcal{F}_{\bf z}=\prod_{t \in \mathbb{Z}_{-}}F(\cdot , {\bf z} _t) \circ p _{t-1}(\mathbf{x}),
\end{equation}
where the projections $p _t: K _L \longrightarrow \overline{B_{\left\|\cdot \right\|}( {\bf 0},L)} $ are given by $p _t({\bf x})= {\bf x}_t $. These projections are continuous when we consider in $K _L $ the product topology. Additionally, the continuity of the reservoir $F$ implies that $\mathcal{F}_{\bf z}  $ is a product of continuous functions, which ensures that $\mathcal{F}_{\bf z} $ is itself continuous  \cite[Theorem 19.6]{Munkres:topology}. Moreover, by the corollaries \ref{all weighted norms are the same} and \ref{km compact complete}, the space $K _L $ is a compact and convex subset of the Banach space $\left(\ell ^{w}_-({\Bbb R}^n), \| \cdot \| _w\right) $ (see Proposition \ref{lw is a banach space}), for any weighting sequence $w$. Schauder's Fixed Point Theorem (see \cite[Theorem 7.1, page 75]{Shapiro:Farrago}) guarantees then that $\mathcal{F}_{\bf z} $ has at least a fixed point, that is, a point $\mathbf{x} \in K _L$ that satisfies $\mathcal{F}_{\bf z} (\mathbf{x})= \mathbf{x}$ or, equivalently,
\begin{equation*}
\mathbf{x}_t=F(\mathbf{x}_{t-1}, {\bf z}_t), \quad \mbox{for all $t \in \Bbb Z_-$},
\end{equation*}
which implies that $\mathbf{x} $ is a solution of $F $ for ${\bf z} $, as required.

\medskip

\noindent {\bf Proof of part (ii)} The main tool in the proof of this part is a parameter dependent version  of the Contraction Fixed Point Theorem, that we include here for the sake of completeness and whose proof can be found in \cite[Theorem 6.4.1, page 137]{Sternberg:dynamical:book}.

\medskip

\noindent {\bf Lemma} {\it 
Let $(X, d_X)$ be a complete metric space and let $Z$ be a metric space. Let $K:X \times Z \longrightarrow X $ be a continuous map such that  for each $z \in Z $, the map $K  _z:X \longrightarrow X $ given by $K _z(x):=K(x,z)$ is a contraction with a constant $0<r<1 $ (independent of $z$), that is, $d_X(K(x,z), K(y,z))\leq r d(x,y) $, for all $x,y \in X $ and all $z \in Z $. Then:
\begin{description}
\item [(i)] For each $z \in Z $, the map $K _z  $ has a unique fixed point in $X$.
\item [(ii)] The map $U _K:Z \longrightarrow X $ that associates to each point $z \in Z $  the unique fixed point of $K _z $ is continuous.
\end{description}
}

\medskip

\noindent Consider now the map 
\begin{equation*}
\begin{array}{cccc}
\mathcal{F}: & K _L \times K _M&\longrightarrow &K _L\\
	&(\mathbf{x}, {\bf z})&\longmapsto & \left(\mathcal{F}(\mathbf{x}, {\bf z})\right)_t:=F(\mathbf{x} _{t-1}, {\bf z}_t).
\end{array}
\end{equation*}
First, as we did in \eqref{product decomposition}, it is easy to show that $\mathcal{F} $ is continuous with respect to the product topologies in $K _M $ and $K _L $, by writing it down as the product of the composition of continuous functions. Second, we show that the map $\mathcal{F} $ is a contraction. Indeed, since by Corollary \ref{all weighted norms are the same} we can choose an arbitrary weighting sequence  to generate the product topologies in $K _M $ and $K _L $, we select $w: \mathbb{N} \longrightarrow (0, 1]$ given by $w _t:= \lambda ^t $, with $t \in \mathbb{N} $  and $\lambda >0$  that satisfies $ 0<r< \lambda<1  $. Then, for any $\mathbf{x}, {\bf y} \in K _L $ and any ${\bf z} \in K _M $, we have
\begin{equation*}
\left\|\mathcal{F}(\mathbf{x}, {\bf z})-\mathcal{F}(\mathbf{y}, {\bf z})\right\|_w=
\sup_{t \in \mathbb{Z}_{-}}\left\{\left\|F(\mathbf{x}_{t-1}, {\bf z}_t)-F(\mathbf{y}_{t-1}, {\bf z}_t)\right\|\lambda^{-t}\right\}\leq 
\sup_{t \in \mathbb{Z}_{-}}\left\{\left\|\mathbf{x}_{t-1}-\mathbf{y}_{t-1}\right\|r\lambda^{-t}\right\},
\end{equation*}
where we used that $F$ is a contraction. Now, since $ 0<r< \lambda<1  $ and hence $r/ \lambda<1 $, we have
\begin{equation*}
\sup_{t \in \mathbb{Z}_{-}}\left\{\left\|\mathbf{x}_{t-1}-\mathbf{y}_{t-1}\right\|r\lambda^{-t}\right\}=
\sup_{t \in \mathbb{Z}_{-}}\left\{\left\|\mathbf{x}_{t-1}-\mathbf{y}_{t-1}\right\|\lambda^{-(t-1)}\frac{r}{\lambda}\right\}\leq
\frac{r}{\lambda} \left\|\mathbf{x}- {\bf y}\right\|_w.
\end{equation*}
This shows that $\mathcal{F} $ is a family of contractions with constant $r/ \lambda<1 $ that is continuously parametrized by the elements in $K _M$. The lemma above implies the existence of a continuous map $U _F: \left(K _M, \left\|\cdot \right\|_w\right)\longrightarrow\left(K _L, \left\|\cdot \right\|_w\right) $ that is uniquely determined by the identity
\begin{equation*}
\mathcal{F} \left(U _F({\bf z}), {\bf z}\right)=U _F({\bf z}), \quad \mbox{for all ${\bf z}\in K _M $}.
\end{equation*}
Proposition \ref{esp implies ti} implies that $U _F $ is causal and time-invariant.  The set $U _F (K _M)$ of accessible states  of the filter $U _F  $ is compact because it is the image of a compact set (see Corollary \ref{km compact complete}) by a continuous map (see \cite[Theorem 26.5, page 166]{Munkres:topology}).

\medskip

\noindent {\bf Proof of part (iii)} Let ${\bf z} \in K _M$ and let $U_{F _1}({\bf z}) $ be the unique solution for ${\bf z} $ of the reservoir systems associated to $F _1 $ available by the part {\bf (ii)} of the theorem that we just proved. Additionally, let  $U_{F _2}({\bf z}) $ be the value of a generalized filter associated to $F _2 $ that exist by hypothesis. Then, for any $t \in \mathbb{Z}_{-} $, we have:
\begin{align*}
\|U_{F _1}({\bf z})_t&-U_{F _2}({\bf z})_t\| = \left\|F _1(U_{F _1}({\bf z})_{t-1}, {\bf z} _t)-F _2(U_{F _2}({\bf z})_{t-1}, {\bf z} _t)\right\|\\
&= \left\|F _1(U_{F _1}({\bf z})_{t-1}, {\bf z} _t)-F _1(U_{F _2}({\bf z})_{t-1}, {\bf z} _t)+F _1(U_{F _2}({\bf z})_{t-1}, {\bf z} _t)-F _2(U_{F _2}({\bf z})_{t-1}, {\bf z} _t)\right\|\\
&\leq \left\|F _1(U_{F _1}({\bf z})_{t-1}, {\bf z} _t)-F _1(U_{F _2}({\bf z})_{t-1}, {\bf z} _t)\right\|+\left\|F _1(U_{F _2}({\bf z})_{t-1}, {\bf z} _t)-F _2(U_{F _2}({\bf z})_{t-1}, {\bf z} _t)\right\|\\
&\leq r\left\|U_{F _1}({\bf z})_{t-1}-U_{F _2}({\bf z})_{t-1}\right\|+\left\|F _1(U_{F _2}({\bf z})_{t-1}, {\bf z} _t)-F _2(U_{F _2}({\bf z})_{t-1}, {\bf z} _t)\right\|.
\end{align*}
If we now recursively apply $n$ times the same procedure to the first summand of this expression, we obtain that  
\begin{multline}
\label{decomposition ineqs}
\|U_{F _1}({\bf z})_t-U_{F _2}({\bf z})_t\| \leq r ^n \|U_{F _1}({\bf z})_{t-n}-U_{F _2}({\bf z})_{t-n}\|+\left\|F _1(U_{F _2}({\bf z})_{t-1}, {\bf z} _t)-F _2(U_{F _2}({\bf z})_{t-1}, {\bf z} _t)\right\|\\
	+r\left\|F _1(U_{F _2}({\bf z})_{t-2}, {\bf z} _{t-1})-F _2(U_{F _2}({\bf z})_{t-2}, {\bf z} _{t-1})\right\|\\
	+ \cdots+r^{n-1}\left\|F _1(U_{F _2}({\bf z})_{t-n}, {\bf z} _{t-(n+1)})-F _2(U_{F _2}({\bf z})_{t-n}, {\bf z} _{t-(n+1)})\right\|
\end{multline}
If we combine the inequality \eqref{decomposition ineqs} with the hypothesis
\begin{equation*}
\left\|F _1-F _2\right\|_{\infty}=\sup_{\mathbf{x} \in \overline{B_{\left\|\cdot \right\|}({\bf 0}, L)},\,  \mathbf{z} \in \overline{B_{\left\|\cdot \right\|}({\bf 0}, M)}} \left\{\left\|F _1(\mathbf{x}, {\bf z})-F _2(\mathbf{x}, {\bf z})\right\|\right\}< \delta(\epsilon):=(1-r) \epsilon,
\end{equation*}
we obtain
\begin{multline}
\label{decomposition ineqs second}
\|U_{F _1}({\bf z})-U_{F _2}({\bf z})\| _\infty = \sup _{t \in \mathbb{Z}_{-}} \left\{
\|U_{F _1}({\bf z})_t-U_{F _2}({\bf z})_t\|  
\right\}\\
\leq 2 L r ^n +(1+ \cdots + r^{n-1}) \delta(\epsilon)=2 L r ^n +\frac{1-r ^n}{1-r} \delta(\epsilon)
\end{multline}
Since this inequality is valid for any $n \in \mathbb{N} $, we can take the limit $n \longrightarrow \infty $ and we obtain that
\begin{equation*}
\|U_{F _1}({\bf z})-U_{F _2}({\bf z})\| _\infty\leq \frac{\delta(\epsilon)}{1-r}= \epsilon.
\end{equation*}
Additionally, as this relation is valid for any ${\bf z} \in K _M $, we can conclude that
\begin{equation*}
\vertiii{U_{F _1}-U_{F _2}}_{\infty}=\sup_{{\bf z} \in K _M} \left\{\left\|U_{F _1} ({\bf z})-U_{F _2} ({\bf z})\right\|_\infty\right\}\leq \epsilon,
\end{equation*}
as required. \quad $\blacksquare$

\medskip

As a straightforward corollary of the first part of the previous theorem, it is easy to show that echo state networks always have (generalized) reservoir filters associated as well as to formulate conditions that ensure simultaneously the echo state and the fading memory properties.

We recall that a map $\sigma: \mathbb{R} \longrightarrow [-1,1] $ is a {\bfi  squashing function} if it is non-decreasing, $\lim_{x \rightarrow -\infty} \sigma(x)=-1 $, and $\lim_{x \rightarrow \infty} \sigma(x)=1 $.

\begin{corollary}
\label{esns have filters}
Consider echo state network given by
\begin{empheq}[left={\empheqlbrace}]{align}
\mathbf{x} _t &=\sigma \left(A\mathbf{x}_{t-1}+ C{\bf z} _t+ \boldsymbol{\zeta}\right),\label{esn reservoir equation theorem prep}\\
{\bf y} _t &= {W}\mathbf{x} _t, \label{esn readout theorem prep}
\end{empheq}
where $C \in \mathbb{M}_{N, n} $ for some $N \in \mathbb{N} $, $\boldsymbol{\zeta} \in \mathbb{R} ^N $,  $A \in \mathbb{M}_{N,N}$,  $W \in \mathbb{M}_{d, N} $, and the input signal ${\bf z} \in \left(D_n\right)^{\mathbb{Z}}$, with $D_n  \subset \mathbb{R}^n$ a compact and convex subset.
The function $\sigma :\mathbb{R}^N \longrightarrow [-1,1] ^N$ in \eqref{esn reservoir equation theorem prep} is constructed by componentwise application of a squashing function that we also call $\sigma$. Then:
\begin{description}
\item [(i)] If the squashing function $\sigma  $ is continuous, then  the reservoir equation \eqref{esn reservoir equation theorem prep} has the existence of solutions property and we can hence associate to the system \eqref{esn reservoir equation theorem prep}-\eqref{esn readout theorem prep} a generalized reservoir filter.
\item [(ii)] If the squashing function $\sigma  $ is differentiable with Lipschitz constant $L _\sigma:=\sup_{x \in \mathbb{R}}\{| \sigma' (x)|\}  < \infty$ and the matrix $A$ is such that $\left\|A\right\|_2 L _\sigma= \sigma_{{\rm max}}(A) L _\sigma<1$, then the reservoir system \eqref{esn reservoir equation theorem prep}-\eqref{esn readout theorem prep} has the echo state and the fading memory properties and we can hence associate to it a unique time-invariant reservoir filter.
\end{description}
The statement in part {\bf  (i)} remains valid when $[-1,1] ^N$ is replaced by a compact and convex subset $D_N \subset [-1,1] ^N$ that is left invariant by the reservoir equation \eqref{esn reservoir equation theorem prep}, that is, $ \sigma \left(A\mathbf{x} + C{\bf z} + \boldsymbol{\zeta}\right) \in D_N$ for any $\mathbf{x} \in D_N$ and any ${\bf z} \in D _n$. The same applies to part {\bf  (ii)} but only the compactness hypothesis is necessary.
\end{corollary}

\begin{remark}
\normalfont
The hypothesis $\left\|A\right\|_2 L _\sigma<1 $ appears in the literature as a sufficient condition to ensure the echo state property, which has been extensively studied in the ESN literature~\cite{jaeger2001, Jaeger04, Buehner:ESN, zhang:echo, Yildiz2012, Wainrib2016, Manjunath:Jaeger}. Our result shows that this condition 
implies automatically the fading memory property. Nevertheless, that condition is far from being sharp and has been significantly improved in \cite{Buehner:ESN, Yildiz2012}. We point out that the enhanced sufficient conditions for the echo state property contained in those references also imply the fading memory property via part {\bf (ii)} of Theorem \ref{uniform approx theorem}.
\end{remark}

\section{Echo state networks as universal uniform approximants}
\label{Echo state networks as universal uniform approximants}

The internal approximation property that we introduced in part {\bf (ii)} of Theorem \ref{uniform approx theorem} tells us that we can approximate any reservoir filter by finding an approximant for the reservoir system that generates it. This reduces the problem of proving a density statement in a space of operators between infinite-dimensional spaces to a space of functions with finite dimensional variables and values. This topic is the subject of many results in approximation theory, some of which we mentioned in the introduction. This strategy allows one to find  simple approximating reservoir filters for any reservoir system that has the fading memory property. In the next result we use as approximating family  the echo state networks that we presented in the introduction and that, as we  see later on, are the natural generalizations of neural networks in a dynamic learning setup, with the important added feature that they are constructed using linear readouts. The combination of this approach with a previously obtained result \cite{RC6} on the density of reservoir filters on the fading memory category allows us to prove in the next theorem that echo state networks can approximate any fading memory filter. On other words, {\it echo state networks are universal}. 

All along this section, we  use the Euclidean norm for the finite dimensional spaces, that is, for each $\mathbf{x}\in \mathbb{R} ^n $, we  write $\left\|\mathbf{x}\right\|:=\left(\sum_{i=1}^n x _i^2 \right)^{1/2} $. For any $M>0 $, the symbol $B_{\left\|\cdot \right\|} ({\bf 0}, M)$ (respectively $\overline{B_{\left\|\cdot \right\|} ({\bf 0}, M)}$) denotes here the open (respectively closed) balls with respect to that norm. Additionally, we set $I _n:=B_{\left\|\cdot \right\|} ({\bf 0}, 1)$. 
\begin{theorem}
\label{ESN universality theorem}
Let  $U:I _n^{\mathbb{Z}_{-}} \longrightarrow \left(\mathbb{R}^d\right)^{\mathbb{Z}_{-}} $ be a causal and time-invariant filter that has the fading memory property. Then, for any $\epsilon>0 $ and any weighting sequence $w$, there is an echo state network 
\begin{empheq}[left={\empheqlbrace}]{align}
\mathbf{x} _t &=\sigma \left(A\mathbf{x}_{t-1}+ C{\bf z} _t+ \boldsymbol{\zeta}\right),\label{esn reservoir equation theorem}\\
{\bf y} _t &= {W}\mathbf{x} _t. \label{esn readout theorem}
\end{empheq}
whose associated generalized filters  $U_{{\rm ESN}}:I _n^{\mathbb{Z}_{-}} \longrightarrow \left(\mathbb{R}^d\right)^{\mathbb{Z}_{-}} $ satisfy that 
\begin{equation}
\label{esn approx}
\vertiii{U-U_{{\rm ESN}}}_{\infty}< \epsilon.
\end{equation}
In these  expressions $C \in \mathbb{M}_{N, n} $ for some $N \in \mathbb{N} $, $\boldsymbol{\zeta} \in \mathbb{R} ^N $,  $A \in \mathbb{M}_{N,N}$, and $W \in \mathbb{M}_{d, N} $.
The function $\sigma :\mathbb{R}^N \longrightarrow [-1,1]^N$ in \eqref{esn reservoir equation theorem} is constructed by componentwise application of a continuous squashing function $\sigma:\mathbb{R} \longrightarrow [-1,1]$ that we denote with the same symbol. 

When the approximating echo state network \eqref{esn reservoir equation theorem}-\eqref{esn readout theorem} satisfies the echo state property, then it has a unique filter $U_{{\rm ESN}} $ associated which is necessarily time-invariant. The corresponding  reservoir functional  $H_{{\rm ESN}}:I _n^{\mathbb{Z}_{-}} \longrightarrow  \mathbb{R}^d  $ satisfies that
\begin{equation}
\label{esn approx functional}
\vertiii{H _U-H_{{\rm ESN}}}_{\infty}< \epsilon.
\end{equation}
\end{theorem}

\begin{remark}
\normalfont
Echo state networks are generally used in practice in the following way: the architecture parameters $A$, $C$, and $\boldsymbol{\zeta}$ are drawn at random from a given distribution and it is only the readout matrix $W$ that is trained using a teaching signal by solving a linear regression problem. It is important to emphasize that the universality theorem that we just stated does not completely explain the empirically observed robustness of ESNs with respect to the choice of those parameters. In the context of standard feedforward neural networks this feature has been addressed using, for example, the so called extreme learning machines \cite{Huang2006}. In dynamical setups and for ESNs this question remains an open problem that will be addressed in future works. 

\end{remark}

\noindent\textbf{Proof of the theorem.\ \ }As we already explained, we proceed by first approximating the filter $U$ by one of the non-homogeneous state-affine system (SAS) reservoir filters introduced in \cite{RC6}, and we later on show that we can approximate that reservoir filter by an echo state network  like the one in \eqref{esn reservoir equation theorem}-\eqref{esn readout theorem}.

We start by recalling that a non-homogeneous state-affine system is a reservoir system determined by the state-space transformation:
\begin{empheq}[left={\empheqlbrace}]{align}
\mathbf{x} _t &=p({\bf z _t})\mathbf{x}_{t-1}+q( {{\bf z}} _t),\label{sas reservoir equation rc7}\\
{\bf y} _t &= W_1\mathbf{x} _t, \label{sas readout rc7}
\end{empheq}
where the inputs ${\bf z _t} \in I _n:=B_{\left\|\cdot \right\|} ({\bf 0}, 1)$, the states $\mathbf{x} _t \in \mathbb{R}^{N _1}$, for some $N _1 \in \mathbb{N} $, and $W _1\in \mathbb{M}_{d,N _1}$. The symbols $p({\bf z _t}) $ and $q({\bf z _t})$ stand for polynomials with matrix coefficients and degrees $r$ and $s$, respectively, of the form:
\begin{eqnarray*}
p({\bf z})&=&\sum_{{i _1, \ldots, i _n \in \left\{0, \ldots, r\right\} \above 0 pt i _1+ \cdots + i _n\leq r}}
z _1^{i _1} \cdots z _n^{i _n} A_{{i _1, \ldots, i _n}}, \quad A_{{i _1, \ldots, i _n}} \in \mathbb{M}_{N_1}, \quad {\bf z} \in I _n\\
q({\bf z})&=&\sum_{{i _1, \ldots, i _n \in \left\{0, \ldots, s\right\} \above 0 pt i _1+ \cdots + i _n\leq s}}
z _1^{i _1} \cdots z _n^{i _n} B_{{i _1, \ldots, i _{n}}}, \quad B_{{i _1, \ldots, i _n}} \in \mathbb{M}_{N_1,1}, \quad {\bf z} \in I _n.
\end{eqnarray*}
Let $L>0$ and choose a real number $K$ such that  
\begin{equation}
\label{condition on k}
0<K< \frac{L}{L+1}<1. 
\end{equation}
Consider now SAS filters that satisfy that $\max _{{\bf z} \in I_n}\sigma_{{\rm max}}(p ({\bf z}))<K $ and $ \max _{{\bf z} \in I_n}\sigma_{{\rm max}}(q ({\bf z})) <K $. It can be shown \cite[Proposition 3.7]{RC6} that under those hypotheses, the reservoir system~\eqref{sas reservoir equation rc7}-\eqref{sas readout rc7} has the echo state property and  defines a unique causal, time-invariant, and fading memory filter $U_{{W_{1}}}^{p,q}:I_n^{\Bbb Z_-} \longrightarrow (\mathbb{R} ^d)^{\Bbb Z_-}  $. Moreover, Theorem 3.12 in \cite{RC6} shows that for any $\epsilon_1>0 $, there exists a SAS filter $U_{{W_{1}}}^{p,q} $ satisfying the hypotheses that we just discussed,  for which
\begin{equation}
\label{first approximation U by functionals}
\vertiii{H_U-H_{{W_{1}}}^{p,q}}_{\infty}< \epsilon  _1,
\end{equation}
where $H_U$ and $H_{{W_{1}}}^{p,q}$ are the reservoir functionals associated to $U$ and $U_{{ W_{1}}}^{p,q} $, respectively. Proposition \ref{linear homeomorphism prop} together with this inequality imply that
\begin{equation}
\label{first approximation U}
\vertiii{U-U_{{W_{1}}}^{p,q}}_{\infty}< \epsilon  _1.
\end{equation}

We now show that the SAS filter $U_{{W_{1}}}^{p,q} $ can be approximated by the filters generated by an echo state network. Define the map
\begin{equation}
\label{sas first version}
\begin{array}{cccc}
F_{{\rm SAS}}: &\overline{B_{\left\|\cdot \right\|} ({\bf 0}, L)} \times I _n &\longrightarrow &\mathbb{R}^{N _1} \\
	&(\mathbf{x}, {\bf z})&\longmapsto & p({\bf z })\mathbf{x}+q( {{\bf z}}),
\end{array}
\end{equation}
with $B_{\left\|\cdot \right\|} ({\bf 0}, L) \subset \mathbb{R}^{N _1} $ and  $p$ and $q$ the polynomials associated to the approximating SAS filter $U_{{W_{1}}}^{p,q} $ in \eqref{first approximation U}. 

The prescription on the choice of the constant $K$ in \eqref{condition on k} has two main consequences. Firstly, the map $F_{{\rm SAS}} $ is a contraction. Indeed, for any $(\mathbf{x}, {\bf z}), (\mathbf{y}, {\bf z}) \in \overline{B_{\left\|\cdot \right\|} ({\bf 0}, L)} \times I _n $:
\begin{equation}
\label{fsas contraction}
\left\|F_{{\rm SAS}}(\mathbf{x}, {\bf z})- F_{{\rm SAS}}(\mathbf{y}, {\bf z})\right\|\leq \left\|p({\bf z })\mathbf{x}-p({\bf z })\mathbf{y}  \right\|\leq \left\|p({\bf z})\right\|_2 \left\|\mathbf{x}- {\bf y}\right\|\leq K\left\|\mathbf{x}- {\bf y}\right\|.
\end{equation}
The map $F_{{\rm SAS}} $ is hence a contraction since $K<1  $ by hypothesis. Second, $\left\|F_{{\rm SAS}}\right\|_ \infty<L $ because by \eqref{condition on k}
\begin{equation*}
\left\|F_{{\rm SAS}}\right\|_ \infty=\sup_{(\mathbf{x}, {\bf z})\in B_{\left\|\cdot \right\|} ({\bf 0}, L) \times I _n}\{ \left\|p({\bf z })\mathbf{x}+q( {{\bf z}})\right\|\}\leq
\sup_{(\mathbf{x}, {\bf z})\in B_{\left\|\cdot \right\|} ({\bf 0}, L) \times I _n} \{\left\|p({\bf z })\right\|_2
\left\|\mathbf{x}\right\|
+\left\|q( {{\bf z}})\right\|\}\leq KL+K<L.
\end{equation*}
This implies, in particular, that the map $F_{{\rm SAS}} $ maps into $\overline{B_{\left\|\cdot \right\|} ({\bf 0}, L)} $ and hence \eqref{sas first version} can be rewritten as
\begin{equation*}
F_{{\rm SAS}}: \overline{B_{\left\|\cdot \right\|} ({\bf 0}, L)} \times I _n \longrightarrow \overline{B_{\left\|\cdot \right\|} ({\bf 0}, L)}.
\end{equation*}
Additionally, we set
\begin{equation}
\label{def of l_1}
L _1:=\left\|F_{{\rm SAS}}\right\|_ \infty<L.
\end{equation}

The uniform density on compacta of the family of feedforward neural networks with one hidden layer proved in \cite{cybenko, hornik} guarantees that for any $\epsilon_2>0 $, there exists $N  \in \mathbb{N}$, $G \in \mathbb{M}_{N, N_1} $, $C \in \mathbb{M}_{N ,n} $, $E \in \mathbb{M}_{N_1, N} $, and $\boldsymbol{\zeta} \in {\Bbb R}^N $, such that  the map defined by
\begin{equation}
\label{NN first version}
\begin{array}{cccc}
F_{{\rm NN}}: &\overline{B_{\left\|\cdot \right\|} ({\bf 0}, L)} \times I _n &\longrightarrow &\mathbb{R}^{N _1} \\
	&(\mathbf{x}, {\bf z})&\longmapsto & E \sigma \left(G \mathbf{x}+ C {\bf z}+ \boldsymbol{\zeta}\right),
\end{array}
\end{equation}
satisfies that
\begin{equation}
\label{approx of f by fnn}
\left\|F_{{\rm NN}} -F _{{\rm SAS}}\right\|_{\infty}=\sup_{\mathbf{x} \in B_{\left\|\cdot \right\|} ({\bf 0}, L),\,  \mathbf{z} \in I _n} \left\{\left\|F_{{\rm NN}} (\mathbf{x}, {\bf z})-F _{{\rm SAS}}(\mathbf{x}, {\bf z})\right\|\right\}< \epsilon _2.
\end{equation}
The combination of \eqref{approx of f by fnn} with the reverse triangle inequality implies that $\left\|F_{{\rm NN}}\right\|_{\infty} -\left\|F _{{\rm SAS}}\right\|_{\infty}< \epsilon _2 $ or, equivalently,
\begin{equation}
\label{fnn and fsas epsilon}
\left\|F_{{\rm NN}}\right\|_{\infty} <\left\|F _{{\rm SAS}}\right\|_{\infty}+ \epsilon _2.
\end{equation}
Given that $\left\|F_{{\rm SAS}}\right\|_ \infty=L _1<L $, if we choose $\epsilon _2>0 $ small enough so that 
$L _1+ \epsilon _2< L $ or, equivalently, 
\begin{equation}
\label{condition on epsilon2}
\epsilon _2< L- L _1,
\end{equation}
then \eqref{fnn and fsas epsilon} guarantees that $\left\|F_{{\rm NN}}\right\|_ \infty<L $, which shows that $F_{{\rm NN}} $ maps into $B_{\left\|\cdot \right\|} ({\bf 0}, L) $, that is, we can write that 
\begin{equation}
\label{fnn maps to ball}
F_{{\rm NN}}: \overline{B_{\left\|\cdot \right\|} ({\bf 0}, L)} \times I _n \longrightarrow \overline{B_{\left\|\cdot \right\|} ({\bf 0}, L)}.
\end{equation}
The continuity of the map $F _{{\rm NN}}$ and the first part of Theorem \ref{uniform approx theorem} imply that the corresponding reservoir equation has the existence of solutions property and that we can hence associate to it a (generalized) filter $U_{F _{{\rm NN}}} $. At the same time, as we proved in \eqref{fsas contraction}, the map $F_{{\rm SAS}}$ is a contraction with constant $K<1$. These  facts, together with \eqref{approx of f by fnn} and the  internal approximation property in Theorem \ref{uniform approx theorem} allow us to conclude that the (unique) reservoir filter $U_{F _{{\rm SAS}}} $ associated to the reservoir map $F _{{\rm SAS}} $ is such that 
\begin{equation}
\label{eps with 2 nana}
\vertiii{U_{F_{{\rm NN}}} -U_{F _{{\rm SAS}}}}_{\infty}< \epsilon _2/(1-K).
\end{equation}
Consider now the readout map $h_{W _1}: \mathbb{R}^{N _1}\longrightarrow \mathbb{R}^d $ given by $h_{W _1}(\mathbf{x}):=W _1 \mathbf{x} $ and let $U_{F_{{\rm NN}}}^{h_{W _1}}:(I _n) ^{\mathbb{Z}_{-}} \longrightarrow (\mathbb{R} ^d) ^{\mathbb{Z}_{-}} $ be the filter given by $U_{F_{{\rm NN}}}^{h_{W _1}}({\bf z})_t:=W _1U_{F_{{\rm NN}}}({\bf z})_t  $, $t \in \mathbb{Z}_{-} $. Analogously, define $U_{F_{{\rm SAS}}}^{h_{W _1}}:(I _n) ^{\mathbb{Z}_{-}} \longrightarrow (\mathbb{R} ^d) ^{\mathbb{Z}_{-}} $ and notice that $U_{F_{{\rm SAS}}}^{h_{W _1}}=U^{p,q}_{W _1} $. Using these observations and \eqref{eps with 2 nana} we have proved that for any $\epsilon _2> 0$ we can find a filter of the type $U_{F_{{\rm NN}}}^{h_{W _1}} $ that satisfies that
\begin{equation}
\label{eps with 2 nana2}
\vertiii{U^{p,q}_{W _1}-U_{F_{{\rm NN}}}^{h_{W _1}} }_{\infty}\leq \left\|W _1\right\|_2 \vertiii{ U_{F _{{\rm SAS}}}- U_{F_{{\rm NN}}}}_{\infty}<  \left\|W _1\right\|_2\epsilon _2/(1-K).
\end{equation}
Consequently, for any $\epsilon>0 $, if we first set $\epsilon_1= \epsilon/2 $ in \eqref{first approximation U by functionals} and we then choose
\begin{equation}
\label{epsilon2 22}
\epsilon_2:=\min \left\{\frac{\epsilon(1-K) }{2\left\|W _1\right\|_2}, \frac{L-L _1}{2}\right\},
\end{equation}
in view of \eqref{condition on epsilon2} and \eqref{eps with 2 nana2}, we can guarantee using \eqref{first approximation U} and \eqref{eps with 2 nana2} that
\begin{equation}
\label{epsilon approx by NSNs}
\vertiii{U-U_{F_{{\rm NN}}}^{h_{W _1}} }_{\infty}\leq \vertiii{U-U^{p,q}_{W _1} }_{\infty} + \vertiii{U^{p,q}_{W _1}-U_{F_{{\rm NN}}}^{h_{W _1}} }_{\infty}\leq \frac{\epsilon}{2}+\frac{\epsilon}{2}= \epsilon.
\end{equation}
In order to conclude the proof it suffices to show that the  filter $U_{F_{{\rm NN}}}^{h_{W _1}} $ can be realized as the reservoir filter associated to an echo state network of the type presented in the statement. We carry that out by using the elements that appeared in the construction of the reservoir $F_{{\rm NN}} $ in \eqref{NN first version} to define a new reservoir map $F_{{\rm ESN}} $ with the architecture of an echo state network. Let $A:=GE \in \mathbb{M}_N $ and define 
\begin{equation}
\label{ESN first version}
\begin{array}{cccc}
F_{{\rm ESN}}: &D _N \times I _n &\longrightarrow &\mathbb{R}^{N } \\
	&(\mathbf{x}, {\bf z})&\longmapsto &  \sigma \left(A \mathbf{x}+ C {\bf z}+ \boldsymbol{\zeta}\right).
\end{array}
\end{equation}
The set $D_N $ in the domain of $F_{{\rm ESN}} $ is given by
\begin{equation}
\label{domain of ESN}
D_N:=[-1,1]^N\cap E ^{-1}(\overline{B_{\left\|\cdot \right\|} ({\bf 0}, L)}),
\end{equation}
where $E ^{-1}(\overline{B_{\left\|\cdot \right\|} ({\bf 0}, L)}) $ denotes the preimage of the set $\overline{B_{\left\|\cdot \right\|} ({\bf 0}, L)} \subset \mathbb{R}^{N _1} $ by the linear map $E: \mathbb{R}^N \longrightarrow \mathbb{R}^{N _1} $ associated to the matrix $E \in \mathbb{M}_{N_1, N} $. This set is compact as $E ^{-1}(\overline{B_{\left\|\cdot \right\|} ({\bf 0}, L)}) $ is closed and $[-1,1]^N  $ is compact and hence $D_N $ is a closed subspace of a compact space which is always compact \cite[Theorem 26.2]{Munkres:topology}. Additionally, $D_N $ is also convex because $[-1,1]^N  $ is convex and $E ^{-1}(\overline{B_{\left\|\cdot \right\|} ({\bf 0}, L)})$ is also convex because it is the preimage of a convex set by a linear map, which is always convex.

We note now that the image of $F_{{\rm ESN}} $ is contained in $D _N $. First, as the squashing function maps into the interval $[-1,1]$, it is clear that 
\begin{equation}
\label{step 1 esn inclusion}
F_{{\rm ESN}} \left(D_N, I _n\right)\subset [-1,1] ^N.
\end{equation}
Second, for any $\mathbf{x} \in D _N  $ we have by construction that $\mathbf{x} \in E ^{-1}(\overline{B_{\left\|\cdot \right\|} ({\bf 0}, L)}) $ and hence $E \mathbf{x} \in \overline{B_{\left\|\cdot \right\|} ({\bf 0}, L)} $. Since by \eqref{fnn maps to ball}  $F _{{\rm NN}} $ maps into $\overline{B_{\left\|\cdot \right\|} ({\bf 0}, L)} $, we can ensure that for any ${\bf z} \in I _n $, the image $F _{{\rm NN}}(E \mathbf{x}, {\bf z}) =E \sigma \left(GE \mathbf{x}+ C {\bf z}+ \boldsymbol{\zeta}\right)=E \sigma \left(A \mathbf{x}+ C {\bf z}+ \boldsymbol{\zeta}\right) \in \overline{B_{\left\|\cdot \right\|} ({\bf 0}, L)} $ or, equivalently, 
\begin{equation}
\label{step 2 esn inclusion}
F_{{\rm ESN}}(\mathbf{x}, {\bf z})=\sigma \left(A \mathbf{x}+ C {\bf z}+ \boldsymbol{\zeta}\right) \in E ^{-1}(\overline{B_{\left\|\cdot \right\|} ({\bf 0}, L)}).
\end{equation}
The relations \eqref{step 1 esn inclusion} and \eqref{step 2 esn inclusion} imply that 
\begin{equation}
\label{step 3 esn inclusion}
F_{{\rm ESN}} \left(D_N, I _n\right)\subset D_N,
\end{equation}
and hence, we can rewrite \eqref{ESN first version} as  
\begin{equation*}
F_{{\rm ESN}} :D _N \times I _n \longrightarrow D _N.
\end{equation*}
The continuity of the map $F _{{\rm ESN}}$ and the compactness and convexity of the set $D_N  \subset \mathbb{R}^N$ that we established above allow us to use the first part of Theorem \ref{uniform approx theorem} to conclude that the corresponding reservoir equation  has the existence of solutions property and that we can hence associate to it a (generalized) filter $U_{F _{{\rm ESN}}} $. Let $W:=W _1E \in \mathbb{M}_{ d,n}$ and define the readout map $h_{{\rm ESN}}:D _N \longrightarrow \mathbb{R}^d $ by $h_{{\rm ESN}}(\mathbf{x}):=W \mathbf{x}= W _1E \mathbf{x} $. Denote by $U _{{\rm ESN} } $ any generalized reservoir filter associated to the echo state network system  $\left(F_{{\rm ESN}},h_{{\rm ESN}}\right) $ that, by construction, satisfies $U_{{\rm ESN}}({\bf z})_t:= h_{{\rm ESN}}(U_{F _{{\rm ESN}}}({\bf z})_t)=WU_{F _{{\rm ESN}}}({\bf z})_t$, for any ${\bf z} \in I _n$ and $t \in \Bbb Z_- $.

We next show that the map $f: D_N=[-1,1]^N\cap E ^{-1}(\overline{B_{\left\|\cdot \right\|} ({\bf 0}, L)}) \longrightarrow\overline{B_{\left\|\cdot \right\|} ({\bf 0}, L)} $ given by $f (\mathbf{x}):= E \mathbf{x} $ is a morphism between the echo state network system $\left(F_{{\rm ESN}},h_{{\rm ESN}}\right) $ and the reservoir system $\left(F_{{\rm NN}},h_{W _1}\right) $. Indeed, the reservoir equivariance property holds because, for any $(\mathbf{x}, {\bf z}) \in D _N \times I _n $, the definitions \eqref{NN first version} and \eqref{ESN first version} ensure that
\begin{equation*}
f(F_{{\rm ESN}}(\mathbf{x}, {\bf z}))=E\sigma \left(A \mathbf{x}+ C {\bf z}+ \boldsymbol{\zeta}\right)=E\sigma \left(GE \mathbf{x}+ C {\bf z}+ \boldsymbol{\zeta}\right)=F_{{\rm NN}}(E\mathbf{x}, {\bf z})=F_{{\rm NN}}(f(\mathbf{x}), {\bf z}).
\end{equation*}
The readout invariance is obvious. This fact and the second part in Proposition \ref{morphisms consequences} imply that all the generalized filters $U _{{\rm ESN} } $ associated to the echo state network are actually filters generated by the system $\left(F_{{\rm NN}},h_{W _1}\right) $. This means that for each generalized  filter $U _{{\rm ESN} } $ there exists a generalized filter of the type $U_{F_{{\rm NN}}}^{h_{W _1}} $ such that $U _{{\rm ESN} }=U_{F_{{\rm NN}}}^{h_{W _1}} $. The inequality \eqref{epsilon approx by NSNs} proves then \eqref{esn approx} in the statement of the theorem. The last claim in the theorem is a straightforward consequence of Propositions \ref{esp implies ti} and \ref{linear homeomorphism prop}.\quad $\blacksquare$

\section{Appendices}

\subsection{Proof of Proposition \ref{esp implies ti}}
\label{proof of lemma esp implies ti}

Let  $\tau \in \mathbb{N} $ and let $T_\tau^n:(D_n) ^{\Bbb Z} \longrightarrow(D_n) ^{\Bbb Z} $ and $T_\tau^N:(D_N) ^{\Bbb Z} \longrightarrow(D_N) ^{\Bbb Z} $ be the corresponding time delay operators. For any ${\bf z} \in (D_n) ^{\Bbb Z} $, let ${\bf x} \in (D_N) ^{\Bbb Z} $ be the unique solution of the reservoir system determined by $F$, that is, 
\begin{equation}
\label{ttau 1}
\mathbf{x}:=U ^F({\bf z}).
\end{equation}
Then, for any $t \in \Bbb Z $,
\begin{equation}
\label{ttau 2}
\left( T _\tau^N \circ U ^F({\bf z}) \right)= \mathbf{x}_{t- \tau}.
\end{equation}
Analogously, let $\widetilde{ \mathbf{x}} \in (D_N) ^{\Bbb Z} $ be the unique solution of $F$ associated to the input $T _\tau ^n({\bf z}) $, that is,
\begin{equation}
\label{ttau 3}
\widetilde{\mathbf{x}} _t=\left( U ^F\circ T _\tau ^n({\bf z}) \right)_t, \quad \mbox{for any} \quad t \in \Bbb Z.
\end{equation}
By construction, the sequence $\widetilde{\mathbf{x}} $ satisfies that
\begin{equation*}
\widetilde{\mathbf{x}} _t= F \left(\widetilde{\mathbf{x}} _{t-1},T _\tau ^n({\bf z})_t\right)= F \left(\widetilde{\mathbf{x}} _{t-1},{\bf z}_{t- \tau}\right), \quad \mbox{for any} \quad t \in \Bbb Z.
\end{equation*}
It we set $s:=t - \tau $, this expression can be rewritten as
\begin{equation}
\label{ttau 4}
\widetilde{\mathbf{x}} _{s+ \tau}=  F \left(\widetilde{\mathbf{x}} _{s+ \tau-1},{\bf z}_{s}\right), \quad \mbox{for any} \quad s \in \Bbb Z,
\end{equation}
and if we define $\widehat{\mathbf{x}} _s:= \widetilde{\mathbf{x}} _{s+ \tau} $, the equality  \eqref{ttau 4} becomes
\begin{equation*}
\widehat{\mathbf{x}} _s=F \left(\widehat{\mathbf{x}} _{s-1}, {\bf z} _s \right), \quad \mbox{for any} \quad s \in \Bbb Z,
\end{equation*}
which shows that $\widehat{\mathbf{x}} \in (D_N) ^{\Bbb Z}$ is a solution of $F$ determined by the input ${\bf z}\in (D_n) ^{\Bbb Z} $. Since the sequence $\mathbf{x} \in (D_N) ^{\Bbb Z}$ in \eqref{ttau 1} is also a solution of $F$ for the same input, the echo state property hypothesis on the systems determined by $F$ implies that $\mathbf{x}= \widehat{\mathbf{x}} $, necessarily. This implies that $\mathbf{x}_{t- \tau}= \widehat{\mathbf{x}}_{t- \tau} $ for all $t \in \Bbb Z $, which is equivalent to $\widetilde{\mathbf{x}} _t= \mathbf{x}_{t- \tau}$. This equality guarantees that \eqref{ttau 2} and \eqref{ttau 3} are equal and since $\widehat{\mathbf{z}} \in (D_n) ^{\Bbb Z}$ is arbitrary, we have that
\begin{equation*}
T _\tau^N \circ U ^F=U ^F\circ T _\tau ^n,
\end{equation*}
as required. \quad $\blacksquare$

\subsection{Proof of Proposition \ref{continuous functional iff filter}}

Suppose first that $U$ is continuous. This implies the existence of a positive function $\delta_U (\epsilon) $ such that if $\mathbf{u},\mathbf{v} \in \left(D_n\right)^{\mathbb{Z}_-} $ are such that $\left\|\mathbf{u}- \mathbf{v}\right\|_{\infty} <\delta _U(\epsilon)$, then $\left\|U(\mathbf{u})-U(\mathbf{v})\right\|_{\infty}< \epsilon $. Under that hypothesis, it is clear that:
\begin{equation*}
\left\| H _U(\mathbf{u})-H _U(\mathbf{v})\right\|= \left\|U(\mathbf{u})_0-U(\mathbf{v})_0\right\|\leq \sup _{t \in \mathbb{Z}_{-} } \left\{ \left\|U(\mathbf{u})_t-U(\mathbf{v})_t\right\|\right\}=\left\|U(\mathbf{u})-U(\mathbf{v})\right\|_{\infty}< \epsilon,
\end{equation*}
which shows the continuity of $H_U : \left(\left(D_n\right)^{\mathbb{Z}_-}, \left\|\cdot \right\|_{\infty} \right) \longrightarrow \left(D_N,  \left\|\cdot \right\|\right) $.

Conversely, suppose that $H : \left(\left(D_n\right)^{\mathbb{Z}_-}, \left\|\cdot \right\|_{\infty} \right) \longrightarrow \left(D_N,  \left\|\cdot \right\|\right) $  is continuous and let $\delta_{H} (\epsilon)>0 $ be such that if $\left\|\mathbf{u}- \mathbf{v}\right\|_{\infty} <\delta _{H}(\epsilon)$ then $\left\| H (\mathbf{u})-H (\mathbf{v})\right\|< \epsilon$. Then, for any $t \in \mathbb{Z}_{-} $,
\begin{equation}
\label{intermediate hu}
 \left\|U_H(\mathbf{u})_t-U_H(\mathbf{v})_t\right\|=
 \left\| H ((\mathbb{P} _{\mathbb{Z}_{-}}\circ T_{-t})(\mathbf{u}))-H ((\mathbb{P} _{\mathbb{Z}_{-}}\circ T_{-t})(\mathbf{v}))\right\|< \epsilon,
\end{equation}
which proves the continuity of $U_H$.
The  inequality follows from the fact that for any $\mathbf{u}\in \left(D_n\right)^{\mathbb{Z}_-} $, the components of the sequence $(\mathbb{P} _{\mathbb{Z}_{-}}\circ T_{-t})(\mathbf{u}) $ are included in those of $\mathbf{u} $ and hence $\sup _{s \in \Bbb Z_-} \left\{\left\|(\mathbb{P} _{\mathbb{Z}_{-}}\circ T_{-t}(\mathbf{u}))_s\right\|\right\} \leq \sup _{s \in \Bbb Z_-} \left\{\left\|\mathbf{u}_s\right\|\right\}$ or, equivalently, $\left\|(\mathbb{P} _{\mathbb{Z}_{-}}\circ T_{-t})(\mathbf{u})\right\|_{\infty}\leq \left\|\mathbf{u}\right\|_{\infty} $. This implies that if $\left\|\mathbf{u}- \mathbf{v}\right\|_{\infty} <\delta _{H}(\epsilon)$ then $\left\|T_{-t}(\mathbf{u})- T_{-t}(\mathbf{v})\right\|_{\infty} <\delta _{H}(\epsilon)$ and hence \eqref{intermediate hu} holds. \quad $\blacksquare$

\subsection{Proof of Theorem \ref{product topology for uniformly bounded}}
\label{proof of product topology for uniformly bounded}

We first show that the map  $D_w^M:({\Bbb R}^n)^{\mathbb{Z}_{-}} \times ({\Bbb R}^n)^{\mathbb{Z}_{-}} \longrightarrow [0, \infty) $ defined in \eqref{def of weighted metric} is indeed a metric. It is clear that $D_w^M(\mathbf{x}, {\bf y})\geq 0 $ and that $D_w^M(\mathbf{x}, {\bf x})= 0 $, for any $\mathbf{x}, {\bf y} \in ({\Bbb R}^n)^{\mathbb{Z}_{-}} $. Conversely, if $D_w^M(\mathbf{x}, {\bf y})= 0 $, this implies that $\overline{d} _M(\mathbf{x}_t, {\bf y}_t)w_{-t}\leq \sup_{t \in \mathbb{Z}_{-}} \left\{\overline{d} _M(\mathbf{x}_t, {\bf y}_t)w_{-t}\right\} =D_w^M(\mathbf{x}, {\bf y})=0$, which ensures that $\overline{d} _M(\mathbf{x}_t, {\bf y}_t)=0 $, for any $t \in \mathbb{Z}_{-} $, and hence $\mathbf{x}= {\bf y} $ necessarily since the map $\overline{d} _M$ is a metric in $\mathbb{R}^n$~\cite[Chapter 2, \textsection{20}]{Munkres:topology}. It is also obvious that $D_w^M(\mathbf{x}, {\bf y})=D_w^M(\mathbf{y}, {\bf x}) $. Regarding the triangle inequality, notice that for any $\mathbf{x}, {\bf y}, {\bf z} \in ({\Bbb R}^n)^{\mathbb{Z}_{-}} $ and $t \in \mathbb{Z}_{-}  $:
\begin{equation*}
\overline{d} _M(\mathbf{x}_t, {\bf z}_t)w_{-t}\leq \overline{d} _M(\mathbf{x}_t, {\bf y}_t)w_{-t}+\overline{d} _M(\mathbf{y}_t, {\bf z}_t)w_{-t}\leq D_w^M(\mathbf{x}, {\bf y})+D_w^M(\mathbf{y}, {\bf z}),
\end{equation*}
which implies that 
\begin{equation*}
D_w^M(\mathbf{x}, {\bf z})=\sup_{t \in \mathbb{Z}_{-}} \left\{\overline{d} _M(\mathbf{x}_t, {\bf z}_t)w_{-t}\right\}\leq D_w^M(\mathbf{x}, {\bf y})+D_w^M(\mathbf{y}, {\bf z}).
\end{equation*}

We now show that the metric topology on $(\mathbb{R})^{\mathbb{Z}_{-}} $ associated to $D_w^M $ coincides with the product topology. Let ${\bf x} \in ({\Bbb R}^n)^{\mathbb{Z}_{-}} $ and let $B_{D_w^M}(\mathbf{x}, \epsilon) $ be an $\epsilon $-ball around it with respect to the metric $D_w^M $. Let now $N \in \mathbb{N}  $  be large enough so that $w _N < \epsilon/M $. We then show that the basis element $V$ for the product topology in  $(\mathbb{R}^n)^{\mathbb{Z}_{-}} $ given by 
\begin{equation*}
V:= \cdots \times  {\Bbb R}^n\times {\Bbb R}^n\times B_{\overline{d} _M}(\mathbf{x}_{-N}, \epsilon) \times \cdots B_{\overline{d} _M}(\mathbf{x}_{-1}, \epsilon) \times B_{\overline{d} _M}(\mathbf{x}_{0}, \epsilon)
\end{equation*}
and that obviously contains the element $\mathbf{x}\in (\mathbb{R}^n)^{\mathbb{Z}_{-}} $ is such that $V \subset B_{D_w^M}(\mathbf{x}, \epsilon) $. Indeed, since for any $ {\bf y}\in (\mathbb{R}^n)^{\mathbb{Z}_{-}} $ and any $t \in \mathbb{Z}_{-}  $ we have that $\overline{d} _M(\mathbf{x} _t, {\bf y}_t)\leq M $, we can conclude that
\begin{equation*}
\overline{d} _M(\mathbf{x} _t, {\bf y}_t)w_{-t}\leq M w_{N}, \quad \mbox{for all} \quad t\leq -N.
\end{equation*} 
Therefore, $D_w^M(\mathbf{x}, {\bf y}) \leq \max \left\{M w_{-N}, \overline{d} _M(\mathbf{x} _{-N}, {\bf y}_{-N})w_{N}, \ldots, \overline{d} _M(\mathbf{x} _{-1}, {\bf y}_{-1})w_{1}, \overline{d} _M(\mathbf{x} _{0}, {\bf y}_{0})w_{0}\right\}$ and hence if
${\bf y}\in V  $ this expression is smaller than $\epsilon $ which allows us to conclude the desired inclusion $V \subset B_{D_w^M}(\mathbf{x}, \epsilon) $.

Conversely, consider a basis element of the product topology given by $U=\prod_{t \in \mathbb{Z}_{-}}U _t $ where $U _t=B_{\overline{d} _M}(\mathbf{x}_{t}, \epsilon_t) $ for a finite set of indices  $t \in \left\{\alpha _1, \ldots, \alpha_r\right\} $, $\epsilon _t\leq 1 $, and $U _t= \mathbb{R}^n $ for the rest. Let $\epsilon := {\rm min}_{t \in \left\{\alpha _1, \ldots, \alpha_r\right\}} \left\{\epsilon _t w _{-t}\right\}$. We now show that $B_{D_w^M}(\mathbf{x}, \epsilon) \subset U$. Indeed, if ${\bf y} \in B_{D_w^M}(\mathbf{x}, \epsilon) $ then $\overline{d} _M(\mathbf{x} _t, {\bf y}_t)w_{-t}\leq D_w^M(\mathbf{x}, {\bf y}) < \epsilon$, for all $t \in \mathbb{Z}_{-} $. It $t \in \left\{\alpha _1, \ldots, \alpha_r\right\} $ then $\epsilon< \epsilon _t w _{-t} $ and hence $\overline{d} _M(\mathbf{x} _t, {\bf y}_t)w_{-t}< \epsilon _t w _{-t} $, which ensures that $\overline{d} _M(\mathbf{x} _t, {\bf y}_t) < \epsilon _t $ and hence ${\bf y} \in U $, as desired.

We conclude by showing that $\left(({\Bbb R}^n)^{\mathbb{Z}_{-}}, D_w^M\right) $ is a complete metric space. First, notice that since for any $\mathbf{x}, {\bf y} \in ({\Bbb R}^n)^{\mathbb{Z}_{-}} $ and any given $t \in \mathbb{Z}_{-} $ we have that 
\begin{equation*}
\overline{d} _M(\mathbf{x}_t, {\bf y}_t)\leq \frac{D_w^M(\mathbf{x}, {\bf y})}{w_{-t}},
\end{equation*}
we can conclude that if $ \left\{\mathbf{x} (i)\right\}_{i \in \mathbb{N}} $ is a Cauchy sequence in $({\Bbb R}^n)^{\mathbb{Z}_{-}} $, then so are the sequences $ \left\{\mathbf{x}_t (i)\right\}_{i \in \mathbb{N}} $ in ${\Bbb R}^n $, for any $t \in \mathbb{Z}_{-}  $, with respect to the bounded metric $\overline{d} _M$. Since the completeness with respect to the bounded metric $\overline{d} _M$ and the Euclidean metric are equivalent~\cite[Chapter 7, \textsection{43}]{Munkres:topology} we can ensure that $ \left\{\mathbf{x}_t (i)\right\}_{i \in \mathbb{N}} $ converges to an element $\mathbf{a} _t \in {\Bbb R}^n $ with respect to the Euclidean metric for any $t \in \mathbb{Z}_{-}  $. We now show that $ \left\{\mathbf{x} (i)\right\}_{i \in \mathbb{N}} $ converges to $\mathbf{a}:= \left(\mathbf{a} _t\right)_{t \in \mathbb{Z}_{-}} \in ({\Bbb R}^n)^{\mathbb{Z}_{-}} $, with respect to the metric $D_w^M$, which proves the completeness statement.

Indeed, since the metric $D_w^M  $ generates the product topology, let $U=\prod_{\in \in \mathbb{Z}_{-}}U _t $ be a basis element such that ${\bf a} \in U $ and, as before, $U _t=B_{\overline{d} _M}(\mathbf{a}_{t}, \epsilon_t) $ for a finite set of indices  $t \in \left\{\alpha _1, \ldots, \alpha_r\right\} $, $\epsilon _t\leq 1 $, and $U _t= \mathbb{R}^n $ for the rest. Let $\epsilon=\min  \left\{\epsilon _{\alpha _1}, \ldots, \epsilon_{\alpha _r}\right\} $. Since for each $t \in \mathbb{Z}_{-} $ the sequence $\mathbf{x}_t (i) \overset{i \rightarrow \infty}{\longrightarrow} \mathbf{a} _t $, then  there exists $N _t \in \mathbb{N}  $ such that for any $k> N _t  $ we have that $\left\|\mathbf{x} _t(k)- \mathbf{a} _t\right\|< \epsilon$.  If we take  $N _\epsilon=\max  \left\{N _{\alpha _1}, \ldots, N_{\alpha _r}\right\} $ then it is clear that $\mathbf{x} (i)  \in U$, for all $i > N _\epsilon $, as required. \quad $\blacksquare$

\subsection{Proof of Corollary \ref{all weighted norms are the same}}
\label{proof of all weighted norms are the same}

Notice first that for any $\mathbf{x}, {\bf y} \in K _M $, we have that $\left\|\mathbf{x}_t- {\bf y}_t\right\|<2M $, $t \in \mathbb{Z}_{-} $, and hence
\begin{equation*}
D_w^{2M}(\mathbf{x}, {\bf y}):=\sup_{t \in \mathbb{Z}_{-}} \left\{\overline{d} _{2M}(\mathbf{x}_t, {\bf y}_t)w_{-t}\right\}=\sup_{t \in \mathbb{Z}_{-}} \left\{\left\|\mathbf{x}_t- {\bf y}_t\right\|w_{-t}\right\}= \left\|\mathbf{x}- {\bf y}\right\|_w.
\end{equation*}
Hence, the topology induced by the weighted norm $\left\|\cdot \right\|_w $ on $K _M $ coincides with the metric topology induced by the restricted metric $D_w^{2M}|_{K _M \times K _M} $ which, by Theorem \ref{product topology for uniformly bounded}, is the subspace topology induced by the product topology on $ \left({\Bbb R}^n\right)^{\mathbb{Z}_{-}} $ on $K _M$  (see \cite[Exercise 1, page 133]{Munkres:topology}), as well as the product topology on the product $K _M=\left(\overline{B_{\left\|\cdot \right\|}(\mathbf{0}, M)}\right)^{\mathbb{Z}_{-}}$ (see \cite[Theorem 19.3, page 116]{Munkres:topology})).\quad $\blacksquare$

\subsection{Proof of Corollary \ref{km compact complete}}

First, since $K _M=\left(\overline{B_{\left\|\cdot \right\|}(\mathbf{0}, M)}\right)^{\mathbb{Z}_{-}} $, it is clearly the product of compact spaces. By Tychonoff's Theorem (\cite[Chapter 5]{Munkres:topology}) $K _M $  is  compact when endowed with the product topology which, by Corollary \ref{all weighted norms are the same}, coincides with the topology associated to the restriction of the norm $\left\|\cdot \right\|_w $ to $K _M $, as well as with the metric topology given by $D^{2M}_w|_{K _M \times K _M} $.

Second, since $ \left(K _M, \left\|\cdot \right\|_w\right) $ is metrizable it is a Hausdorff space. This implies (see \cite[Theorem 26.3]{Munkres:topology}) that as $K _M  $ is a compact subspace of the Banach space $(\ell ^{w}_-({\Bbb R}^n), \left\|\cdot \right\|_w) $ (see Proposition \ref{lw is a banach space}) then it is necessarily closed. This in turn implies  (\cite[Theorem B, page 72]{simmons:topology}) that $ \left(K _M, \left\|\cdot \right\|_w\right) $ is complete. 

Finally, the convexity statement follows from the fact that the product of convex sets is always convex.
\quad $\blacksquare$

\subsection{Proof of Proposition \ref{in lww norm finer than product}}

Let $d_w $ be the metric on $\ell ^{w}_-({\Bbb R}^n)$ induced by the weighted norm $ \left\|\cdot \right\|_w $ and let $D_w:=D_w ^1 $ be the $w$-weighted metric on $({\Bbb R}^n)^{\mathbb{Z}_{-}} $ with constant $M=1 $ introduced in Theorem \ref{product topology for uniformly bounded} and defined using the same underlying norm in ${\Bbb R}^n  $ as the one associated to $\left\|\cdot \right\|_w $. As we saw in that theorem, the metric $D_w$ induces the product topology on $({\Bbb R}^n)^{\mathbb{Z}_{-}} $. 

Let now $\mathbf{u} \in \ell ^{w}_-({\Bbb R}^n)$  and  let $\epsilon >0$. Let now $\mathbf{v} \in \ell ^{w}_-({\Bbb R}^n)$ be such that $d_w(\mathbf{u}, \mathbf{v})< \epsilon $. By definition, we have that
\begin{equation*}
D_w(\mathbf{u}, \mathbf{v})=\sup_{t \in \mathbb{Z}_{-}} \left\{\overline{d} _1(\mathbf{x}_t, {\bf y}_t)w_{-t}\right\}=\sup_{t \in \mathbb{Z}_{-}} \left\{(\min \{\|\mathbf{x}_t- {\bf y}_t\|, 1\})w_{-t}\right\}\leq
\sup_{t \in \mathbb{Z}_{-}} \left\{\|\mathbf{x}_t- {\bf y}_t\|w_{-t}\right\}=d_w(\mathbf{u}, \mathbf{v})< \epsilon,
\end{equation*}
which shows that $B_{d_w}(\mathbf{u}, \epsilon)\subset B_{D_w}(\mathbf{u}, \epsilon) $ and allows us to conclude that the norm topology in $\ell ^{w}_-({\Bbb R}^n) $ is finer than the subspace topology induced by the product topology in $\left(\mathbb{R}^n\right)^{\mathbb{Z}_{-}} $. 

We now show that this inclusion is strict. Since the weighting sequence $w$ converges to zero, there exists an element $t _0 \in \mathbb{Z}_{-}  $ such that $w_{-t _0}< \epsilon/2 $. Let $\lambda > 0  $ arbitrary and define  the element $\mathbf{v} ^\lambda \in ({\Bbb R}^n)^{\mathbb{Z}_{-}} $ by setting $\mathbf{v} ^\lambda_{t _0}:= \lambda \mathbf{u}_{t _0} $ and $\mathbf{v} ^\lambda_{t}:= \lambda \mathbf{u}_{t } $ when $t \neq t _0 $. We now show that $\mathbf{v} ^ \lambda \in B_{D_w}(\mathbf{u}, \epsilon) $ for any $\lambda>0 $. Indeed,
\begin{equation*}
D_w(\mathbf{u}, \mathbf{v}^\lambda)=\min \{| \lambda-1|\|\mathbf{u}_{t_0}\|, 1\}w_{-t _0}\leq
1 \cdot w_{-t _0}< \epsilon/2< \epsilon.
\end{equation*}
At the same time, by definition, 
\begin{equation*}
d_w(\mathbf{u}, \mathbf{v}^\lambda)=| \lambda-1|\|\mathbf{u}_{t_0}\|w_{-t _0}< \infty,
\end{equation*}
which shows that $\mathbf{v}^\lambda \in \ell ^{w}_-({\Bbb R}^n)$. However, since $| \lambda-1|\|\mathbf{u}_{t_0}\|w_{-t _0} $ can be made as large as desired by choosing $\lambda $ big enough, we have proved that for any ball  $B_{d_w}(\mathbf{u}, \epsilon') $, with $\epsilon'>0 $ arbitrary, the ball $B_{D_w}(\mathbf{u}, \epsilon) $ contains always an element in $\ell ^{w}_-({\Bbb R}^n)$ that is not included in  $B_{d_w}(\mathbf{u}, \epsilon') $. This argument allows us to conclude that the norm topology in $\ell ^{w}_-({\Bbb R}^n) $ is strictly finer than the subspace topology induced by the product topology. \quad $\blacksquare$

\subsection{Proof of Lemma \ref{fs and ^sfor w}}

The proof requires the following preparatory lemma that will also be used later on in the proof of Proposition \ref{FMP independent of w}.

\begin{lemma}
\label{ operator is continuous}
Let $M>0 $ and let $w $ be a weighting sequence. Then:
\begin{description}
\item [(i)] The  operator $\mathbb{P} _{\mathbb{Z}_{-}} \circ T _{-t}: (K _M, \left\|\cdot \right\|_w) \longrightarrow (K _M, \left\|\cdot \right\|_w)$ is a continuous map, for any $t \in \mathbb{Z}_{-}$.
\item [(ii)] The projections $p _i: (\ell ^{w}_-(\mathbb{R}^n), \left\|\cdot \right\|_w) \longrightarrow (\mathbb{R}^n, \left\|\cdot \right\|) $, $i \in \mathbb{Z}_{-}$,  given by $p _i({\bf z})= {\bf z}_i $, are continuous.
\end{description}
\end{lemma}

\noindent\textbf{Proof of the lemma.\ \ (i)} We show that this statement is true by characterizing $\mathbb{P} _{\mathbb{Z}_{-}}  \circ T _{-t} $ as a Cartesian product of continuous maps between two product spaces endowed with the product topologies and by using Corollary \ref{all weighted norms are the same}. Indeed, notice first that the projections $p _i: (K _M, \left\|\cdot \right\|_w) \longrightarrow \overline{B_{\left\|\cdot \right\|}( {\bf 0},M)} $  are continuous since by Corollary \ref{all weighted norms are the same} the topology induced on $K _M $ by the weighted norm $\left\|\cdot \right\|_w $ is the product topolopy. Since $\mathbb{P} _{\mathbb{Z}_{-}}  \circ T _{-t} $ can be written as the infinite Cartesian product  of continuous maps $\mathbb{P} _{\mathbb{Z}_{-}}  \circ T _{-t}=\prod _{i=t}^{- \infty} p _i = \left(\ldots,p_{t-2},p_{t-1}, p _t\right)$ it is hence continuous when using the product topology induced by $\left\|\cdot \right\|_w $ (see \cite[Theorem 19.6]{Munkres:topology}). 

\medskip

\noindent {\bf (ii)} Notice first that the projections $p _i: (\ell ^{w}_-(\mathbb{R}^n), \left\|\cdot \right\|_w) \longrightarrow (\mathbb{R}^n, \left\|\cdot \right\|) $ are obviously continuous when we consider in $\ell ^{w}_-(\mathbb{R}^n) $ the subspace topology induced by the product topology in $({\Bbb R}^n)^{\mathbb{Z}_{-}} $. The continuity of $p _i: (\ell ^{w}_-(\mathbb{R}^n), \left\|\cdot \right\|_w) \longrightarrow (\mathbb{R}^n, \left\|\cdot \right\|) $ then follows directly from Proposition \ref{in lww norm finer than product}. $\blacktriangledown $

\medskip

We now proceed with the proof of Lemma \ref{fs and ^sfor w}. Let first $H \in \mathbb{H}_{K _M} ^{w} $. The FMP hypothesis implies that the map $H:(K _M, \left\|\cdot \right\|_w) \longrightarrow (\mathbb{R}^N, \left\|\cdot \right\| )$ is continuous. Given that $K _M $ is compact by Corollary \ref{km compact complete} then so is $H(K _M) \subset \mathbb{R} ^M$. This in turn implies that $H(K _M) $ is closed and bounded \cite[Theorem 27.3]{Munkres:topology} which guarantees the existence of $L>0 $ such that $H(K_M)\subset \overline{B_{\left\|\cdot \right\|}(\mathbf{0}, L)}) $. The map obtained out of $H$ by restriction of its target to $\overline{B_{\left\|\cdot \right\|}(\mathbf{0}, L)}) $ (that we denote with the same symbol) is  also continuous and hence $H \in \mathbb{H}_{K _M, K _L} ^{w} $.

Let now $U:K _M \longrightarrow \ell ^{w}_-(\mathbb{R}^N)  $  in $\mathbb{F}_{K _M} ^{w} $ and consider the composition $p _0 \circ U: K _M \longrightarrow \mathbb{R}^N $. The FMP hypothesis on $U$ and the continuity of $p _0: (\ell ^{w}_-(\mathbb{R}^n), \left\|\cdot \right\|_w) \longrightarrow (\mathbb{R}^n, \left\|\cdot \right\|) $ that we established in the second part of Lemma \ref{ operator is continuous} imply that $p _0 \circ U $ is continuous. This implies, together with the compactness of $K _M $ that we proved in Corollary \ref{km compact complete}, the existence of $L>0 $ such that $p _0 \circ U(K_M)\subset \overline{B_{\left\|\cdot \right\|}(\mathbf{0}, L)}) $. Equivalently, for any ${\bf z} \in K _M $, we have that $U ({\bf z})_0 \in \overline{B_{\left\|\cdot \right\|}(\mathbf{0}, L)})$. Now, since $U$ is by hypothesis time invariant, we have by \eqref{why we can restrict to zminus} that 
\begin{equation*}
U ({\bf z})_t= \left(T_{-t} \left(U({\bf z})\right)\right)_0= U \left(T_{-t}({\bf z})\right)_0  \in \overline{B_{\left\|\cdot \right\|}(\mathbf{0}, L)}), \  t \in \mathbb{Z}_{-}, \ \mbox{since $T_{-t}({\bf z}) \in K _M  $},
\end{equation*}
which proves that  $U (K _M) \subset  K _L $. The map obtained out of $U$ by restriction of its target to $K _L$ (that we denote with the same symbol) is  also continuous since $(K _L, \left\|\cdot \right\|_w)$ is a topological subspace of $(\ell ^{w}_-(\mathbb{R}^n), \left\|\cdot \right\|_w) $ and hence $U \in \mathbb{F}_{K _M, K _L} ^{w} $, as required.

The inclusion $\mathbb{F}_{K _M, K _L} ^{w} \subset \mathbb{F}_{K _M} ^{w} $ (respectively, $\mathbb{H}_{K _M, K _L} ^{w} \subset \mathbb{H}_{K _M} ^{w} $) is a consequence of the continuity of the inclusion map $(K _L, \left\|\cdot \right\|_w)\hookrightarrow (\ell ^{w}_-(\mathbb{R}^n), \left\|\cdot \right\|_w) $ (respectively, $(\overline{B_{\left\|\cdot \right\|}(\mathbf{0}, L)}), \left\|\cdot \right\|)\hookrightarrow (\mathbb{R}^n, \left\|\cdot \right\|)$). \quad $\blacksquare$

\subsection{Proof of Proposition \ref{FMP independent of w}}

\noindent {\bf Proof of part (i)} The FMP of $U$ with respect to the sequence $w$ is, by definition, equivalent to the continuity of the map $U:(K _M, \left\|\cdot \right\|_w)\longrightarrow (K _L, \left\|\cdot \right\|_w) $ (respectively, $H:(K _M, \left\|\cdot \right\|_w)\longrightarrow (\overline{B_{\left\|\cdot \right\|}( {\bf 0},L)}, \left\|\cdot \right\|) $). By Corollary \ref{km compact complete}, this is equivalent to the continuity of these maps when $K _M$ and $K _L $ are endowed with the product topology which is, by the same result, generated by any arbitrary weighting sequence.

Consider now $U:(K _M, \left\|\cdot \right\|_w)\longrightarrow (\ell ^{w}_-({\Bbb R}^n), \left\|\cdot \right\|_w)$ in $\mathbb{F}_{K _M } ^{w} $ (respectively, $H:(K _M, \left\|\cdot \right\|_w)\longrightarrow ({\Bbb R}^n, \left\|\cdot \right\|) $ in $\mathbb{H}_{K _M } ^{w} $). By Lemma \ref{fs and ^sfor w} there exists an $L >0 $ such that $U$ (respectively, $H$) can be considered an element of $\mathbb{F}_{K _M , K _L} ^{w} $ (respectively, $\mathbb{H}_{K _M , K _L } ^{w} $) by restriction of the target. Using the statement that we just proved about the space $\mathbb{F}_{K _M , K _L} ^{w} $ (respectively, $\mathbb{H}_{K _M , K _L } ^{w} $) we can conclude that $U$ (respectively, $H $) has the FMP with respect to any weighting sequence. Since, again by Lemma \ref{fs and ^sfor w}, the inclusion $\mathbb{F}_{K _M, K _L} ^{w} \subset \mathbb{F}_{K _M} ^{w} $ (respectively, $\mathbb{H}_{K _M, K _L} ^{w} \subset \mathbb{H}_{K _M} ^{w} $) holds true for any $M>0 $, and any weighting sequence $w$, we can conclude that $U$ (respectively, $H$) is continuous as an element of $\mathbb{F}_{K _M } ^{w} $ (respectively, $\mathbb{H}_{K _M } ^{w} $) for any weighting sequence $w$, as required.

\medskip

\noindent {\bf Proof of part  (ii)}  First, suppose that $H:(K _M, \left\|\cdot \right\|_w)\longrightarrow (\overline{B_{\left\|\cdot \right\|}( {\bf 0},L)}, \left\|\cdot \right\|) $ has the FMP and that this map is hence continuous. Given that the associated filter  $U _H:(K _M, \left\|\cdot \right\|_w) \longrightarrow (K _L, \left\|\cdot \right\|_w)$ can be written as $U _H= \prod _{t=0}^{- \infty} H \circ \left(\mathbb{P} _{\mathbb{Z}_{-}}  \circ T _{-t}\right) $ we can also conclude that it is continuous. Indeed, by part {\bf (i)} of Lemma \ref{ operator is continuous}, the map $H \circ \left(\mathbb{P} _{\mathbb{Z}_{-}}  \circ T _{-t}\right)  $, $t \in \mathbb{Z}_{-}$, is a composition of continuous functions and it is hence continuous. Additionally, the product  $\prod _{t=0}^{- \infty} H \circ \left(\mathbb{P} _{\mathbb{Z}_{-}}  \circ T _{-t}\right) :(K _M, \left\|\cdot \right\|_w) \longrightarrow (K _L, \left\|\cdot \right\|_w)$ is also continuous because the topology of $(K _L, \left\|\cdot \right\|_w)$ coincides with the product topology by Corollary \ref{all weighted norms are the same} and hence the continuity follows from \cite[Theorem 19.6]{Munkres:topology}, which shows that $U _H $ has the FMP. Conversely, if $U :(K _M, \left\|\cdot \right\|_w) \longrightarrow (K _L, \left\|\cdot \right\|_w)$ has the FMP, so is the case with $H _U=p _0 \circ U: (K _M, \left\|\cdot \right\|_w) \longrightarrow  (\overline{B_{\left\|\cdot \right\|}( {\bf 0},L)}, \left\|\cdot \right\|) $ as it is the composition of two continuous maps. These arguments shows that $\boldsymbol{\Psi}(\mathbb{F}_{K _M, K _L} ^{{\rm FMP}}) \subset  \mathbb{H}_{K _M, K _L} ^{{\rm FMP}}$ and $\boldsymbol{\Phi}(\mathbb{H}_{K _M, K _L} ^{{\rm FMP}} ) \subset  \mathbb{F}_{K _M, K _L} ^{{\rm FMP}}$, and  that the maps $\boldsymbol{\Psi}:\mathbb{F}_{K _M, K _L} ^{{\rm FMP}} \longrightarrow \mathbb{H}_{K _M, K _L} ^{{\rm FMP}}$ and $\boldsymbol{\Phi}:\mathbb{H}_{K _M, K _L} ^{{\rm FMP}} \longrightarrow \mathbb{F}_{K _M, K _L} ^{{\rm FMP}}$ are hence  inverses of each other.

The parallel statement regarding the spaces $\mathbb{F}_{K _M} ^{{\rm FMP}}$ and $  \mathbb{H}_{K _M} ^{{\rm FMP}}$ can be easily established by mimicking the proof of part {\bf (i)} using Lemma \ref{fs and ^sfor w}. 
\quad $\blacksquare$

\subsection{Proof of Proposition \ref{linear homeomorphism prop}}

We start by proving the continuity of $\boldsymbol{\Psi} $ by establishing the inequality \eqref{first ineq psis}. Let $U \in  \mathbb{F}_{K _M} ^{{\rm FMP}}  $. By definition 
\begin{equation}
\label{step 1 for continuity}
\vertiii{\boldsymbol{\Psi}(U)}_{\infty} =\sup_{{\bf z} \in K _M} \left\{\left\|\boldsymbol{\Psi}(U)({\bf z})\right\|\right\}=
\sup_{{\bf z} \in K _M} \left\{\left\|U ({\bf z})_0\right\|\right\}.
\end{equation}
Since we have that
\begin{equation*}
\sup_{{\bf z} \in K _M} \left\{\left\|U ({\bf z})_0\right\|\right\}\leq
\sup_{{\bf z} \in K _M}  \{\sup_{t \in \Bbb Z_-} \{
\left\|U ({\bf z})_t\right\| \}
\}=
\sup_{{\bf z} \in K _M} \left\{
\left\|U ({\bf z})\right\|_\infty\right\}=
\vertiii{U}_ \infty,
\end{equation*}
this shows, together with \eqref{step 1 for continuity}, that 
\begin{equation*}
\vertiii{\boldsymbol{\Psi}(U)}_{\infty}\leq
\vertiii{U}_ \infty,
\end{equation*}
which implies the continuity of $\boldsymbol{\Psi} $. Regarding the inequality \eqref{second ineq psis}, let $H \in \mathbb{H}_{K _M}^{{\rm FMP}} $. We have:
\begin{multline*}
\vertiii{\boldsymbol{\Phi}(H )}_{\infty}=
\sup_{{\bf z} \in K _M} \left\{\sup_{t \in \mathbb{Z}_{-}}\left\{\left\|\boldsymbol{\Phi}(H ) ({\bf z})_t\right\|\right\}\right\}=
\sup_{{\bf z} \in K _M} \left\{\sup_{t \in \mathbb{Z}_{-}}\left\{\left\|H  ((\mathbb{P} _{\mathbb{Z}_{-}}\circ T_{-t})({\bf z}))\right\|\right\}\right\}\\
\leq
\sup_{{\bf z} \in K _M} \left\{\left\|H  ({\bf z}))\right\|_ \infty\right\}=\vertiii{H }_{\infty}, 
\end{multline*}
which proves the continuity of $\boldsymbol{\Phi} $. The inequality is a consequence of the fact that the sequence $(\mathbb{P} _{\mathbb{Z}_{-}}\circ T_{-t})({\bf z}) \in K _M $.  The inequalities \eqref{first ineq psis kl} and \eqref{second ineq psis kl} are proved in a similar fashion.\quad $\blacksquare$

\subsection{Proof of Corollary \ref{esns have filters}}

\noindent\textbf{(i)} Consider the reservoir map $F_{{\rm ESN}}:[-1,1]^N \times D _n \longrightarrow [-1,1]^N $ given by $F_{{\rm ESN}}:= \sigma \left(A\mathbf{x} + C{\bf z} + \boldsymbol{\zeta}\right)$
The statement is a direct consequence of the continuity of $F_{{\rm ESN}} $, the compactness and convexity of $[-1,1]^N  $  and $D _n $,  and of part {\bf (i)} of Theorem \ref{uniform approx theorem}. 

\medskip

\noindent {\bf (ii)} The result follows from part {\bf (ii)} of Theorem \ref{uniform approx theorem} since the hypotheses in the statement imply that the reservoir map $F_{{\rm ESN}} $  is in those circumstances a contraction. Indeed, let $\mathbf{x}, {\bf y} \in [-1,1]^N $  and let $z \in  D _n $, then
\begin{equation*}
\left\|F_{{\rm ESN}} (\mathbf{x}, {\bf z})-F_{{\rm ESN}} (\mathbf{y}, {\bf z})\right\|= \left\|\sigma \left(A\mathbf{x} + C{\bf z} + \boldsymbol{\zeta}\right)-\sigma \left(A\mathbf{y} + C{\bf z} + \boldsymbol{\zeta}\right)\right\|\leq
L _\sigma \left\|A\right\|_2  \left\|\mathbf{x}- {\bf y}\right\|.
\end{equation*}
Since by hypothesis $L _\sigma \left\|A\right\|_2<1 $, we can conclude that $F_{{\rm ESN}} $ is a contraction, as required. The norm $ \left\|\cdot \right\| $ in the previous expression is the Euclidean norm in $\mathbb{R}^N $. The time invariance of the resulting unique fading memory reservoir filter is a consequence of Proposition \ref{esp implies ti}. \quad $\blacksquare$

\subsection{$\left(\ell ^{w}_-({\Bbb R}^n), \| \cdot \| _w\right) $ is a Banach space}
\label{lw is a banach space appendix}

\begin{proposition}
\label{lw is a banach space}
Let $w : \mathbb{N} \longrightarrow (0,1] $ be a weighting sequence and let $\| \cdot \| _w : (\mathbb{R}^n)^{\Bbb Z _{-}}  \longrightarrow  \overline{\mathbb{R}^+} $ be the corresponding weighted norm. Then, the space $\left(\ell ^{w}_-({\Bbb R}^n), \| \cdot \| _w\right) $ defined by
\begin{equation*}
\ell ^{w}_-({\Bbb R}^n):= \left\{{\bf z}\in \left(\mathbb{R}^n\right)^{\mathbb{Z}_{-}}\mid \| {\bf z}\| _w< \infty\right\},
\end{equation*}
endowed with weighted norm  $\| \cdot \| _w $ is a Banach space.
\end{proposition}

\noindent\textbf{Proof.\ \ } We first show that $\ell ^{w}_- ({\Bbb R}^n)$ is a linear subspace of $(\mathbb{R}^n)^{\Bbb Z _{-}} $. Let $ \mathbf{u}, \mathbf{v} \in \ell ^{w}_- ({\Bbb R}^n) $  and let $\lambda \in \mathbb{R} $. Then,
\begin{multline*}
\left\|\mathbf{u}+ \lambda \mathbf{v}\right\|_w= \sup_{t \in \Bbb Z_-} \left\{\left\|\mathbf{u}_t+ \lambda \mathbf{v} _t\right\|w_{-t}\right\}\leq \sup_{t \in \Bbb Z_-} \left\{\left\|\mathbf{u}_t\right\|w_{-t}+ \lambda \left\|\mathbf{v} _t\right\|w_{-t}\right\}\\
\leq 
\sup_{t \in \Bbb Z_-} \left\{\left\|\mathbf{u}_t\right\|w_{-t}\right\}+ \lambda \sup_{t \in \Bbb Z_-} \left\{\left\|\mathbf{v} _t\right\|w_{-t}\right\}= \left\|\mathbf{u}\right\|_w+ \lambda \left\|\mathbf{v}\right\|_w.
\end{multline*}
We now show that this space is complete. Let $\left\{\mathbf{u} (n)\right\}_{n \in \mathbb{N}} \subset \ell ^{w}_- ({\Bbb R}^n)$ be a Cauchy sequence. This implies that for any $\epsilon>0 $, there exists $N(\epsilon) \in \mathbb{N}  $ such that for all $m,n >N(\epsilon) $ we have $\left\|\mathbf{u}(n)- \mathbf{u} (m)\right\|_w< \epsilon $. Hence, for any $t \in \Bbb Z_- $,
\begin{equation}
\label{ineq for t}
\left\|\mathbf{u}_t(n)- \mathbf{u}_t (m)\right\|w_{-t}\leq\sup_{t \in \Bbb Z_-} \left\{\left\|\mathbf{u}_t(n)- \mathbf{u}_t (m)\right\|w_{-t}\right\}= \left\|\mathbf{u}(n)- \mathbf{u} (m)\right\|_w< \epsilon.
\end{equation}
This implies that taking for each fixed $t \in \mathbb{Z}_{-} $ the value $N(\epsilon w_{-t}) $, the sequences $\left\{\mathbf{u}_t(n)\right\}_{n \in \mathbb{N}} $ in $\mathbb{R}^n $ are Cauchy and hence convergent to values $\mathbf{u}_t\in \mathbb{R}^n $. We now show that $\left\{\mathbf{u} (n)\right\}_{n \in \mathbb{N}} $ converges to $\mathbf{u} \in (\mathbb{R}^n)^{\Bbb Z _{-}}$. Using \eqref{ineq for t}, take $N(\epsilon/2) $ so that for all $m,n>N(\epsilon/2) $ and any $t \in \mathbb{Z}_{-} $ one has $\left\|\mathbf{u}_t(n)- \mathbf{u}_t (n)\right\|w_{-t}\leq \epsilon/2 $. If we take the limit $m \rightarrow \infty $ in this inequality, we obtain 
\begin{equation*}
\left\|\mathbf{u}_t(n)- \mathbf{u}_t \right\|w_{-t}\leq \epsilon/2, \quad \mbox{for all $t \in \mathbb{Z}_{-}$.}
\end{equation*}
This implies that
\begin{equation*}
\left\|\mathbf{u}(n)- \mathbf{u} \right\|_w=\sup_{t \in \Bbb Z_-} \left\{\left\|\mathbf{u}_t(n)- \mathbf{u}_t \right\|w_{-t}\right\}\leq \epsilon/2< \epsilon,
\end{equation*}
which proves that $\left\{\mathbf{u} (n)\right\}_{n \in \mathbb{N}} $ converges to $\mathbf{u} $, as required. It remains to be shown that $\mathbf{u} \in \ell ^{w}_-({\Bbb R}^n) $, that is, that $\left\|\mathbf{u}\right\|< \infty $. In order to show that this is indeed the case, let $n \in \mathbb{N} $ be such that $\left\| \mathbf{u}- \mathbf{u}(n)\right\|< \epsilon $. This implies that 
\begin{equation*}
\left\| \mathbf{u}\right\|_w- \left\|\mathbf{u}(n)\right\|_w<|\left\| \mathbf{u}\right\|_w- \left\|\mathbf{u}(n)\right\|_w|< \left\|\mathbf{u}- \mathbf{u}(n) \right\|_w < \epsilon, 
\end{equation*}
and hence $\left\| \mathbf{u}\right\|_w< \left\| \mathbf{u}(n)\right\|_w + \epsilon < \infty $, as required. \quad $\blacksquare$

\medskip

\noindent {\bf Acknowledgments:} We thank Herbert Jaeger and Josef Teichmann  for carefully looking at early versions of this work and for making suggestions that have significantly improved some of our results. We also thank the editor and two anonymous referees whose input has significantly improved the presentation and the contents of the paper. The authors acknowledge partial financial support of the French ANR ``BIPHOPROC" project (ANR-14-OHRI-0002-02) as well as the hospitality of the Centre Interfacultaire Bernoulli of the Ecole Polytechnique F\'ed\'erale de Lausanne during the program ``Stochastic Dynamical Models in Mathematical Finance, Econometrics, and Actuarial Sciences" that made possible the collaboration that lead to some of the results included in this paper. LG acknowledges partial financial support of the Graduate School of Decision Sciences and the Young Scholar Fund AFF of the Universit\"at Konstanz. JPO acknowledges partial financial support  coming from the Research Commission of the Universit\"at Sankt Gallen and the Swiss National Science Foundation (grant number 200021\_175801/1).

\noindent
\addcontentsline{toc}{section}{Bibliography}
\bibliographystyle{wmaainf}
\bibliography{/Users/JPO/Dropbox/Public/GOLibrary}
\end{document}